\documentclass{article}

\usepackage{PRIMEarxiv}

\usepackage[utf8]{inputenc} 
\usepackage[T1]{fontenc}    
\usepackage{hyperref}       
\usepackage{url}            
\usepackage{booktabs}       
\usepackage{amsfonts}       
\usepackage{nicefrac}       
\usepackage{microtype}      
\usepackage{lipsum}
\usepackage{fancyhdr}       
\usepackage{graphicx}       
\graphicspath{{media/}}     
\usepackage{natbib}

\usepackage{balance} 

\usepackage{hyperref}       
\usepackage{amsmath}
\usepackage{mathtools, amsthm}
\usepackage{adjustbox}
\usepackage{tabularx}
\usepackage{multirow}
\usepackage{tcolorbox}
\usepackage[ruled,linesnumbered,boxed]{algorithm2e}

\newtheorem{theorem}{Theorem}[section]

\newtheorem{lemma}[theorem]{Lemma}

\title{On Regret-optimal Cooperative Nonstochastic Multi-armed Bandits
}

\author{
  Jialin Yi \quad  Milan Vojnovi{\'{c}} \\
  Department of Statistics \\
  The London School of Economics and Political Science \\
  London, UK\\
  \texttt{\{j.yi8,  m.vojnovic\}@lse.ac.uk} \\
}

\begin{document}
\maketitle

\begin{abstract}
    We consider the nonstochastic multi-agent multi-armed bandit problem with agents collaborating via a communication network with delays.
    We show a lower bound for individual regret of all agents.
    We show that with suitable regularizers and communication protocols, a collaborative multi-agent \emph{follow-the-regularized-leader} (FTRL) algorithm has an individual regret upper bound that matches the lower bound up to a constant factor when the number of arms is large enough relative to degrees of agents in the communication graph. We also show that an FTRL algorithm with a suitable regularizer is regret optimal with respect to the scaling with the edge-delay parameter. 
    We present numerical experiments validating our theoretical results and demonstrate cases when our algorithms outperform previously proposed algorithms. 
\end{abstract}


\section{Introduction}\label{sec::intro}

Coordinating multiple agents that can communicate with each other to make decisions under uncertainty is a classical problem and has many different applications in computer science \citep{DBLP:books/mk/Lynch96}, game theory \citep{chakravarty_mitra_sarkar_2014} and machine learning \citep{DBLP:conf/nips/LanctotZGLTPSG17}.
We consider the multi-agent version of a multi-armed bandit problem which
is one of the most fundamental decision making problems under uncertainty.
In this problem, a learning agent needs to consider the exploration-exploitation trade-off, i.e. balancing the exploration of various actions in order to learn how much rewarding they are and selecting high-rewarding actions. In the multi-agent version of this problem, multiple agents collaborate with each other trying to maximize their individual cumulative rewards, and the challenge is to design efficient cooperative algorithms under communication constraints.

We consider the nonstochastic (adversarial) multi-armed bandit problem in a cooperative multi-agent setting, with $K\geq 2$ arms and $N\geq 1$ agents. In each time step, each agent selects an arm and then observes the incurred loss corresponding to its selected arm. The losses of arms are according to an arbitrary loss sequence, which is commonly referred to as the nonstochastic or adversarial setting. Each agent observes only the loss of the arm this agent selected in each time step. The agents are allowed to cooperate by exchanging messages, which is constrained by a communication graph $G$ such that any two agents can exchange a message directly between themselves only if they are neighbors in graph $G$. Each exchange of a message over an edge has delay of $d$ time steps. The goal of each agent is to minimize its cumulative loss over a time horizon of $T$ time steps. We study the objective of minimizing the individual regret of agents, i.e. the difference between the expected cumulative loss incurred by an agent and the cumulative loss of the best arm in hindsight. We also study the average regret of all agents.

The multi-agent multi-armed bandit problem formulation that we study captures many systems that use a network of learning agents. For example, in peer-to-peer recommender systems, the agents are users and the arms are products that can be recommended to users \citep{DBLP:journals/jcss/BaragliaDMR13}. The delay corresponds to the time it takes for a message to be transmitted between users. Note that in this application scenario, the number of products (i.e. arms) may be much larger than the number of users (i.e. agents).

\sloppy The collaborative multi-agent multi-armed bandit problem was studied, e.g., in \citet{DBLP:journals/jmlr/Cesa-BianchiGM19} and \citet{DBLP:conf/nips/Bar-OnM19}, where each edge has unit delay. Our setting is more general in allowing for arbitrary delay $d$ per edge.  \citet{DBLP:journals/jmlr/Cesa-BianchiGM19} showed that when each agent selects arms according to a cooperative Exp3 algorithm (Exp3-Coop), the average regret is $O(\sqrt{(\alpha(G)/N +1/K)\log(K)KT})$ for large enough $T$, where $\alpha(G)$ is the independence number of graph $G$. \citet{DBLP:conf/nips/Bar-OnM19} have shown that individual regret of each agent $v$ is $O(\sqrt{(1/|\mathcal{N}(v)|+1/K)\log(K)KT})$ when $T\geq K^2 \log(K)$, where $\mathcal{N}(v)$ is the set of neighbors of agent $v$ and itself in graph $G$. This regret bound is shown to hold for an algorithm where some agents, referred to as center agents, select arms using the Exp3-Coop policy and other agents copy the actions of center agents. These bounds reveal the effect of collaboration on the learning and what graph properties effect the efficiency of learning. However, some fundamental questions still remain. For example, to the best of our knowledge, it is unknown from the previous literature what is the lower bound for this problem. Moreover, it is unknown whether better algorithms can be designed whose regret matches a lower bound under certain conditions.

In this work, we give a regret lower bound for any learning algorithm in which each agent can only communicate with their neighbors. We present a center-based algorithm whose regret upper bound matches the lower bound when the number of arms is large enough. We present an algorithm that has a regret upper bound with $\sqrt{d}$ dependence on the delay per edge, which is optimal. All our regret bounds are parametrized with the delay parameter $d$, which is unlike to \citet{DBLP:journals/jmlr/Cesa-BianchiGM19} and \citet{DBLP:conf/nips/Bar-OnM19} which considered only the special when $d=1$. In what follows we summarise our results in more details.

\subsection{Summary of our contributions}

We show that any algorithm has individual regret for each agent $v$ lower bounded as 
$$
\Omega\left(\sqrt{\left(\frac{1}{|\mathcal{N}(v)|}+\frac{1}{K}\log(K)\right)KT}\right),
$$
when $T \geq K/|\mathcal{N}(v)|$. This implies the average regret lower bound $\Omega(\sqrt{(\delta(G)^2+\log(K)/K)KT})$, when $T \geq \delta(G)^2K$, where $\delta(G)=(1/N)\sum_{v\in \mathcal{V}}(1/\sqrt{|\mathcal{N}(v)|})$. Hence, there is a $\sqrt{\log(K)}$ factor gap between the previously known upper bounds and the lower bound. 

We show an algorithm that guarantees individual regret for each agent $v$ to be 
$$
O\left(\sqrt{\left(\frac{1}{|\mathcal{N}(v)|}+\frac{|\mathcal{N}(C(v))|}{K}\log(K)^2\right)KT}\right),
$$ 
whenever $K\geq \max_{v}|\mathcal{N}(v)|$ and $T \geq \Omega(K\max_{c\in \mathcal{C}} |\mathcal{N}(c)|)$, where $\mathcal{C}$ is the set of center agents and $C(v)$ is the nearest center agent to agent $v$. This regret bound improves the best known upper bound on individual regret when the number of arms is large enough relative to agents' degrees. 
Moreover, we note that the algorithm has optimal regret up to a constant factor when when $|\mathcal{N}(C(v))||\mathcal{N}(v)|\log(K)^2/K = O(1)$, i.e. when the number of arms is large enough.

Our algorithm is based on using a cooperative Follow-the-Regularized-Leader (FTRL) policy with a Tsallis entropy regularizer. This is in contrast to \citet{DBLP:journals/jmlr/Cesa-BianchiGM19,DBLP:conf/nips/Bar-OnM19}, which both use a cooperative Exp3 policy.

Our regret analysis relies on a key new lemma that bounds the change of the action selection strategy of an agent under Tsallis entropy regularization. This result may be of independent interest. 

We also present a decentralized follow-the-regularized-leader algorithm that has regret with optimal dependency on the delay parameter $d$, namely scaling as $\sqrt{d}$. This algorithm uses a hybrid regularizer, which combines an Exp3 type regularizer with a Tsallis entropy regularizer. This algorithm is decentralised with all agents applying the same strategy.

\subsection{Related work}

The multi-armed bandit problem in a multi-agent setting, where agents collaborate with each other subject to some communication constraints, has received considerable attention in recent years.
\citet{DBLP:journals/jcss/AwerbuchK08a} introduced the cooperative nonstochastic multi-armed bandit problem setting where communication is through a public channel (corresponding to a complete graph) and some agents may be dishonest.
\citet{DBLP:conf/cdc/KarPC11} considered a special collaboration network in which only one agent can observe the loss of the selected arm in each time step. 
 \citet{DBLP:conf/icml/SzorenyiBHOJK13} discussed two specific P2P networks in which at each time step, each agent can send messages to only two other agents.
\citet{DBLP:conf/alt/Cesa-BianchiCM20} studied an online learning problem where only a subset of agents play in each time step. They showed that an optimal average regret bound for this problem is $\Theta(\sqrt{\alpha(G)T})$ when the set of agents that play in each time step is chosen randomly, while $\Omega(T)$ bound holds when the set of agents can be chosen arbitrarily in each time step.
\citet{DBLP:journals/ton/KollaJG18}, \citet{DBLP:conf/cdc/LandgrenSL16, DBLP:conf/cdc/LandgrenSL16} and \citet{DBLP:conf/nips/Martinez-RubioK19} considered a setting in which communication is constrained by a communication graph such that any two agents can communicate \emph{instantly} if there is an edge connecting them.

The communication model considered in our paper was introduced by \citet{DBLP:journals/jmlr/Cesa-BianchiGM19}. Here, agents communicate via messages sent over edges of a fixed connected graph and sending a message over an edge incurs a delay of value $d$. \citet{DBLP:journals/jmlr/Cesa-BianchiGM19} considered the case when $d=1$ whereas in this paper, we consider $d\geq 1$.

They proposed an algorithm, referred to as Exp3-Coop, in which each agent constructs loss estimators for each arm using an importance-weighted estimator.
The Exp3-Coop algorithm has an upper bound of $O(\sqrt{(\alpha(G)/N + 1/K)\log(K)KT} + \log(T))$ on the average regret. 

\citet{DBLP:conf/nips/Bar-OnM19} combines the idea of center-based communication from \citet{DBLP:journals/ton/KollaJG18} with the Exp3-Coop algorithm, showing that the center-based Exp3 algorithm has a regret upper bound of 
$O(\sqrt{(1/|\mathcal{N}(v)|+1/K)\log(K)KT})$ for each individual agent when $d=1$. We show that a better regret bound can be guaranteed with respect to the scaling with the number of arms $K$.

Multi-armed bandits with delayed feedback have been studied extensively in the single-agent setting \citep{DBLP:conf/icml/JoulaniGS13, DBLP:conf/nips/ThuneCS19, DBLP:conf/icml/FlaspohlerOCMOO21}. Specifically,
\citet{DBLP:conf/aistats/ZimmertS20} considered a setting in which the agent has no prior knowledge about the delays and showed an optimal regret of $O(\sqrt{KT}+\sqrt{d\log(K)T})$ where $d$ is the average delay over $T$ time steps. We present, in the multi-agent setting, a distributed learning algorithm whose regret upper bound can also achieve this optimal $\sqrt{d}$ dependence on the delay per edge $d$.

We use the \emph{Tsallis entropy} family of regularizers proposed by \citet{tsallis1988possible}. \citet{DBLP:journals/jmlr/ZimmertS21} have shown that an online mirror descent algorithm with a Tsallis entropy regularizer achieves optimal regret for the single-agent bandit problem. We show a distributed learning algorithm for the multi-agent bandit setting, which uses a Tsallis entropy regularizer.

\subsection{Organization of the paper} 

Section~\ref{sec::problem-formulation} provides problem formulation and definitions of notation.
Section~\ref{sec::algorithms-and-regret-upper-bounds} presents our two algorithms and their regret bounds. In Section~\ref{sec::regret-lower-bound}, we present a lower bound on individual regret of each agent.  Section~\ref{sec::numerical-experiements} contains numerical results. Finally, conclusion remarks are given in Section~\ref{sec::conclusion}. Proofs of our results are available in the supplementary material.

\section{Problem formulation}
\label{sec::problem-formulation}

We consider a multi-armed bandit problem with a finite set  $\mathcal{A}=\{1,\ldots, K\}$ of actions (arms) played by $N$ agents. The agents can communicate through a communication network $G = (\mathcal{V}, \mathcal{E})$
where $\mathcal{V}$ is the set of $N$ agents and $\mathcal{E}$ is the set of edges such that $(u,v)\in \mathcal{E}$ if, and only if, agent $u$ can send/receive messages to/from agent $v$. We denote the neighbors of the agent $v$ and itself by the set $\mathcal{N}(v) = \{u\in \mathcal{V}: (u, v)\in \mathcal{E}\}\cup\{v\}$.
Sending a message over edge $e\in \mathcal{E}$ incurs a delay of value $d_e \geq 0$ time steps. We consider the \emph{homogeneous} setting under which $d_e = d\geq 1$ for every edge $e\in E$.
Note that the delayed communication network model in \citet{DBLP:journals/jmlr/Cesa-BianchiGM19} and \citet{DBLP:conf/nips/Bar-OnM19} is restricted to the special case $d=1$. 



At each time step $t = 1,2, \ldots, T$, each agent $v\in \mathcal{V}$ chooses an action $I_t(v)\in \mathcal{A}$ according to distribution $p_t^v$ over $\mathcal{A}$ and then observes the loss value, $\ell_t(I_t(v)) \in [0,1]$. Notice that the loss does not depend on the agent, but only on the time step and the chosen action. Hence, if two agents choose the same action at the same time step, they incur the same loss. We consider the \emph{nonstochastic setting} where the losses are determined by an oblivious adversary, meaning that the losses do not depend on the agent's realized actions. 

At the end of each time step $t$, each agent $v\in \mathcal{V}$ sends a message $S_t(v)$ of size $b_t(v)$ information bits to all its neighbors and after this, each agent $v\in \mathcal{V}$ has messages $\cup_{u\in \mathcal{N}(v)}\{S_{s}(u): s+d = t\}$.

We assume that at each time step $t$, each agent $v$ can send to each of its neighbors a message $S_t(v)=\left\langle v, t, I_{t}(v), \ell_{t}\left(I_{t}(v)\right), p_{t}^{v}\right\rangle$, i.e. the agent id, the time step, the chosen arm id, the instant loss received and the instant action distribution. We denote with $b_t(v)$ the number of information bits to encode $S_t(v)$.
The total communication cost in each time step is $\sum_{v\in \mathcal{V}}\sum_{u:(u,v)\in \mathcal{E}}b_t(u)$ information bits.

The \emph{individual regret} of each agent $v$ is defined as the difference between its expected accumulated loss and the loss of the best action in hindsight, i.e. 
$$
R_{T}^v=\mathbb{E}\left[\sum_{t=1}^{T} \ell_{t}\left(I_{t}(v)\right)\right]-\min _{i \in \mathcal{A}} \sum_{t=1}^{T} \ell_{t}(i).
$$
The \emph{average regret} of $N$ agents is defined as 
$$
R_{T}=\frac{1}{N} \sum_{v \in \mathcal{V}} R_{T}^v.
$$

\paragraph{Additional notation}
We define $\mathcal{P}_{K-1}$ to be the $K-1$ simplex.

Let $\alpha(G)$ be the size of a maximal independent set of graph $G$, where the maximal independent set is the largest subset of nodes such that no two nodes in this set are connected by an edge.

\section{Algorithms and regret upper bounds}
\label{sec::algorithms-and-regret-upper-bounds}

In this section, we propose two collaborative multi-agent bandit algorithms, the center-based cooperative follow-the-regularized-leader (CFTRL) algorithm and the decentralized cooperative follow-the-regularized-leader (DFTRL) algorithm. The first algorithm has optimal regret up to a constant factor when the number of arms is large enough. The second algorith has optimal depence on the delay parameter $d$.

\subsection{A center-based cooperative follow-the-regularized-leader algorithm}

We consider an algorithm where some agents, referred to as \emph{centers}, run a FTRL algorithm, and each other agent copies the action selection distribution from its nearest center. The strategy based on using center agents was proposed in \citet{DBLP:conf/nips/Bar-OnM19}, where agents played the Exp3 strategy instead. These centre agents collaboratively update their strategies by exchanging messages with other agents, and each non-center agent copies the strategy of its nearest center agent. The center agents are selected such that they have a sufficiently large degree, which can be shown to reduce individual regret of center agents. Moreover, the center agents are selected such that each non-center agent is within a small distance to a center agent. 

Let $\mathcal{C} \subseteq \mathcal{V}$ be the set of centers. The set of agents $\mathcal{V}$ is partitioned into disjoint components $\mathcal{V}_c$, $c\in \mathcal{C}$.
Each non-center agent $v$ belongs to a unique component. For each agent $v$, let $C(v)$ denote its center agent,  $c=C(v)$ if and only if $v\in \mathcal{V}_{c}$.  
Let $d(v)$ be the distance between a non-center agent $v$ and its center $C(v)$. The set of centers $\mathcal{C}$ and the partitioning $\{\mathcal{V}_c: c\in \mathcal{C}\}$ are computed according to Algorithms~3 and 4 in \citet{DBLP:conf/nips/Bar-OnM19}. 

Let $\mathcal{J}_t(v) = \{I_t(v^\prime): v^\prime\in \mathcal{N}(v)\}$ be the set of actions chosen by agent $v$ or its neighbors at time step $t$. 
Each center agent $c\in C$ runs a FTRL algorithm with the collaborative importance-weighted loss estimators observable up to time step $t$, 
$$
\hat{L}^{c,obs}_t(i) = \sum_{s=1}^{t-1}\hat{\ell}^{c,obs}_s(i)
$$
and
$$
\hat{\ell}_{t}^{c,obs}(i)= 
\left\{\begin{array}{cl}
     \frac{\ell_{t-d}(i)}{q_{t-d}^c(i)} \mathbb{I}\left\{i\in\mathcal{J}_{t-d}(c)\right\}  & \text { if } t> d \\
     0 & \text { otherwise }
    \end{array}\right.
$$
where 
$$
q_t^c(i) = 1-\prod_{v \in \mathcal{N}(c)}\left(1-p_t^{v}(i)\right)
$$
is the \emph{neighborhood-aggregated importance weight}.

In each time step, the center agents update their action selection distributions according to the FTRL algorithm, i.e.
$$
p_t^c=\operatorname{argmin}_{p\in \mathcal{P}_{K-1}} \left\{ \left\langle p, \hat{L}_{t}^{c, obs}\right\rangle + F_t(p)\right\}.
$$
where $F_t(p)$ is the \emph{Tsallis entropy regularizer}
\citep{DBLP:journals/jmlr/ZimmertS21} with the learning rate $\eta(c)$, 
\begin{equation}
    \label{eq::tsallis-entropy}
    F_t(p) = -2\sum_{i=1}^K\sqrt{p_i} / \eta(c).
\end{equation}

Each non-center agent $v\in \mathcal{V}\backslash C$ selects actions according to the uniform distribution until time step $t>d(v)d$.
Then the non-center agent copies the action selection distribution from its center, i.e.
$p^v_{t} = p^{C(v)}_{t-d(v)d}.$

The details of the CFTRL algorithm are described in Algorithm \ref{algo:CFTRL}.

\begin{algorithm}[t]
\caption{Center-based cooperative FTRL (CFTRL)}\label{algo:CFTRL}
\SetKwInOut{Init}{Initialization}
\SetKwInOut{Input}{Input}
\Input{Tsallis regularizer Eq.~(\ref{eq::tsallis-entropy}), learning rate $\eta(c)$ and the delay $d$.} 
\Init{$\hat{L}_1^{c, obs}(i) = 0$ for all $i\in \mathcal{A}$ and $c\in C$, $p_1^v(i) = 1/K$ for all $i\in \mathcal{A}$ and $v\not\in C$.}
\For{each time step $t=1,2,\dots, T$}{

    Each $c\in \mathcal{C}$ updates $p_t^c= \operatorname{argmin}_{p\in \mathcal{P}_{K-1}}\{ \langle p, \hat{L}_{t}^{c, obs}\rangle + F_t(p)\} 
    $\;
    Each $c\in \mathcal{C}$ chooses $I_t(c)=i$ with probability $p_t^c(i)$ and receives the loss $\ell_t(I_t(c))$\;
    Each $c\in \mathcal{C}$ sends the message $S_t(c) = \left\langle c, t,  I_t(c), \ell_t(I_t(c)), p_t^c\right\rangle$ to all their neighbors\;
    Each $c\in \mathcal{C}$ receives messages $\{S_{t-d}(v): v\in \mathcal{N}(c)\}$ and computes $\hat{L}_{t+1}^{c, obs}$\;
    Each $v\in \mathcal{V}\setminus \mathcal{C}$ updates $p_{t}^v = p_{t-d(v)d}^{C(v)}$ when $t>d(v)d$ and $p_{t}^v = p_{1}^{v}$ otherwise\;
    Each $v\in V\setminus \mathcal{C}$ chooses $I_t(v)=i$ with probability $p_t^v(i)$ and receives $\ell_t(I_t(v))$\;
    Each $v\in \mathcal{V}\setminus \mathcal{C}$ sends $S_t(v) = \left\langle v, t,   I_t(v), \ell_t(I_t(v)), p_t^v\right\rangle$ to all its neighbors.
}
\end{algorithm}

\subsubsection{Individual regret upper bound}

We show an individual regret upper bound for Algorithm~\ref{algo:CFTRL} in the following theorem.

\begin{theorem}\label{thm::CFTRL}

Assume that $K\geq \max_{v\in \mathcal{V}}|\mathcal{N}(v)|$ and $T\geq 36(d+1)^2K\max_{c}|\mathcal{N}(c)|$, and agents follow the CFTRL algorithm with each center agent $c\in \mathcal{C}$ using the learning rate $\eta(c) = \sqrt{|\mathcal{N}(c)| / (3T)}$. Then, the individual regret of each agent $v\in \mathcal{V}$ is bounded as 
$$
R_T^v = O\left(\frac{1}{ \sqrt{|\mathcal{N}(v)|}} \sqrt{KT} + d\log(K)\sqrt{|\mathcal{N}(C(v))|T}\right).
$$
\end{theorem}
The proof of the theorem is provided in the supplementary material.
 
In the following, we provide a proof sketch. The proof relies on two key lemmas which are shown next.

\begin{lemma}
\label{lm:ftrl-coop:individual}
Assume that the delay of each edge is $d\geq 1$, then the individual regret of each center agent $v$ with the regularizer $F_t(p) = \sum_{i=1}^K f_t(p_i)$ satisfies
\begin{eqnarray*}
R_{T}^v & \leq & M + \frac{1}{2} \mathbb{E} \left[\sum_{t=1}^T  \sum_{i\in \mathcal{A}} \frac{1}{q_t^v(i) f_t^{\prime\prime}(p_t^v(i))}\right] \\
&& + d \cdot \mathbb{E} \left[\sum_{t=1}^T \sum_{i\in \mathcal{A}} \frac{1}{f_t^{\prime\prime}(p_t^v(i))}\right]
\end{eqnarray*}
where 
$$
M = \max_{x \in \mathcal{P}_{K-1}}-F_{1}(x)+\sum_{t=2}^{T} \max_{x \in \mathcal{P}_{K-1}}\left(F_{t-1}(x)-F_{t}(x)\right).
$$
\end{lemma}
\begin{lemma}
\label{lm:ftrl-tsallis:selection-prob}

For any $\delta>1$, assume that agent $v$ runs a FTRL algorithm
with Tsallis entropy (\ref{eq::tsallis-entropy}) and learning rate $\eta(v)\leq (1-1/\sqrt{\delta})/(\delta^{3d/2}\sqrt{K})$ and $K\geq 2$, then for all $t\geq 1$ and $i\in \mathcal{A}$
$$
(1-(1+\delta)\eta(v)\hat{\ell}_t^{v,obs}(i)) p_t^v(i) \leq p_{t+1}^v(i) \leq \delta p_{t}^v(i).
$$
\end{lemma}
The proofs of the two lemmas are provided in the supplementary material. 
Similar property as in Lemma~\ref{lm:ftrl-tsallis:selection-prob} was known  to hold for the Exp3 algorithm by a result in \citet{DBLP:journals/jmlr/Cesa-BianchiGM19}. To the best of our knowledge, this property was previously not known to hold for the FTRL algorithm with Tsallis entropy.

Lemma~\ref{lm:ftrl-coop:individual} bounds the individual regret by the sum of a constant, the regret due to the \emph{instant} bandit feedback, and the regret due to the \emph{delayed} full-information feedback, which is $O(\sqrt{KT/|\mathcal{N}(c)|})$.
Since the action selection distributions of non-center agents are copied from their centers in the past rounds, Lemma~\ref{lm:ftrl-tsallis:selection-prob} bounds the difference between action selection distributions of non-center agent $v$ and its center $C(v)$ in the same rounds when $T>d(v)d$.
Consequently, the difference between the individual regret of a non-center agent $v$ and its center $C(v)$ is bounded by $O(d(v)d \eta(C(v)) T) = O(d\log(K) \sqrt{|\mathcal{N}(C(v))|T})$.

\subsection{A decentralized cooperative follow-the-regularizer-leader algorithm}

Theorem~\ref{thm::CFTRL} provides a bound for individual regrets, which increases linearly in the edge-delay parameter $d$. This can be problematic when the delay in the communication network is large.
We show that the effect of delays on regret can be reduced by using a decentralized follow-the-regularized-leader (DFTRL) algorithm.

In the DFTRL algorithm, each agent runs a FTRL algorithm 
 
with
a \emph{hybrid regularizer} $F_t(p)$ defined in \citet{DBLP:conf/aistats/ZimmertS20} as follows
\begin{equation}
\label{eq::hybrid-regularizer}
F_t(p) = \sum_{i=1}^{K} \left( \frac{- 2  \sqrt{p_{i}}}{\eta_t} + \frac{p_i \log(p_i)}{\zeta_t} \right)
\end{equation}
where $\eta_t$ and $\zeta_t$ are some non-increasing sequences.

As is shown in Theorem~4.1, there is a regret lower bound that consists of two parts: the first part is the regret lower bound of the multi-armed bandit problem and the second part is the regret lower bound of the bandit problem with full-information but delayed feedback \cite{DBLP:conf/icml/JoulaniGS13}. 
The hybrid regularizer combines the Tsallis entropy regularizer with an optimal regularizer in the full-information setting, the negative entropy regularizer.
The learning rates of the two regularizers can be tuned separately to minimize the regret from the two parts.

The details of the DFTRL are described in Algorithm~\ref{algo:DFTRL}.
\begin{algorithm}[t]
\caption{Decentralized cooperative FTRL (DFTRL)}\label{algo:DFTRL}
\SetKwInOut{Init}{Initialization}
\SetKwInOut{Input}{Input}
\Input{Hybrid regularizer Eq.~(\ref{eq::hybrid-regularizer}), learning rates $\eta_t$, $\zeta_t$, and delay $d$.}
\Init{$\hat{L}_1^{v, obs}(i) = 0$ for all $i\in \mathcal{A}$ and $v\in \mathcal{V}$.}
\For{each time step $t=1,2,\dots, T$}{
    Each $v\in \mathcal{V}$ updates $
    p_t^v=\operatorname{argmin}_{p\in \mathcal{P}_{K-1}}\{\langle p, \hat{L}_{t}^{v, obs}\rangle + F_t(p)\}
    $\;
    Each $v\in \mathcal{V}$ chooses $I_t(v)=i$ with probability $p_t^v(i)$ and receives the loss $\ell_t(I_t(v))$\;
    Each $v\in \mathcal{V}$ sends the message $S_t(v) = \left\langle v, t,  I_t(v), \ell_t(I_t(v)), p_t^v\right\rangle$ to all their neighbors\;
    Each $v\in \mathcal{V}$ receives messages $\{S_{t-d}(v^\prime): v^\prime\in \mathcal{N}(v)\}$ and computes $\hat{L}_{t+1}^{v, obs}$\;
}
\end{algorithm}

\subsubsection{Average regret upper bound}

We show a bound on the average regret for the DFTRL algorithm in the following theorem. The sequences $\eta_t$ and $\zeta_t$ are assumed to be set as
$$
\eta_t = (1/(1-1/e)) (\alpha(G)/ N + 1/K)^{-1/4} \sqrt{2/T}
$$
and 
$$
\zeta_t = \sqrt{\log(K) / (dt)}.
$$

\begin{theorem}\label{thm::DFTRL} 

Assume that each agent follows the DFTRL algorithm

and the delay of each edge is $d\geq 1$, then the average regret over $N$ agents is bounded as

$$
R_{T} = O\left( \left(\frac{\alpha(G)}{N} + \frac{1}{K}\right)^{1/4}\sqrt{KT} +  \sqrt{d\log(K)T} \right).
$$
\end{theorem}

Proof of the theorem is provided in the supplementary material. 
We note that the average regret scales as $\sqrt{d}$ which is better than linear scaling of the CFTRL algorithm.

For the special case when $d=1$, as in  \citet{DBLP:journals/jmlr/Cesa-BianchiGM19},Theorem~\ref{thm::DFTRL}
shows that when the number of arms $K$ is large enough, then the DFTRL algorithm has an $O((\alpha(G)/N)^{1/4}\sqrt{KT})$ regret, which is better than $O((\alpha(G)/N)^{1/2}\sqrt{KT}\sqrt{\log(K)})$ of Exp3-Coop from \citet{DBLP:journals/jmlr/Cesa-BianchiGM19}. Specifically, this improvement holds when
$
K = \Omega(\exp(\sqrt{N/\alpha(G)})).
$

In what follows we provide a proof sketch of Theorem~\ref{thm::DFTRL}.
First we present a key lemma whose proof is provided in the supplementary material. 
\begin{lemma}
\label{lm:ftrl-coop-independence}
For every agent $v\in \mathcal{A}$ and any probability distribution $p^v$ over $\mathcal{A}$, it holds
$$
\sum_{i\in \mathcal{A}}\sum_{v \in \mathcal{V}} \frac{p^v(i)^{\frac{3}{2}}}{q^v(i)} \leq 
N\sqrt{\frac{1}{1-1/e}\left(\frac{\alpha(G)}{N}+\frac{1}{K}\right)K}
$$
where $q^v(i) = 1-\prod_{v^\prime \in \mathcal{N}(v)}(1-p^{v^\prime}(i))$.
\end{lemma}

Lemma~\ref{lm:ftrl-coop-independence} shows that the average regret from the FTRL algorithm with the hybrid regularizer with instant feedback is $O((\alpha(G)/N + 1/K)^{1/4}\sqrt{KT})$.
For the regularizer in Eq.~(\ref{eq::hybrid-regularizer}), the delay effect term is $O(\sqrt{dT\log(K)})$.
Lemma~\ref{lm:ftrl-coop:individual} bounds the average regret by the sum of two terms.

\section{Regret lower bounds}
\label{sec::regret-lower-bound}

We present lower bounds on individual regret $R_T^v$ for every agent $v\in \mathcal{V}$ and average regret $R_T$. 

\begin{theorem} \label{thm:lower-2} 
The worst-case individual regret of each agent $v\in \mathcal{V}$, $R_T^v$, is
$$
\Omega\left(\max\left\{ \min \left\{T, \frac{1}{\sqrt{|\mathcal{N}(v)|}}\sqrt{KT}\right\}, \sqrt{d \log(K)T}\right\} \right)
$$
and the worst-case average regret, $R_T$, is
$$
\Omega\left( \max\left\{ \min \left\{T, c_G\sqrt{KT}\right\}, \sqrt{d \log(K)T}\right\} \right)
$$
where $c_G = (1/N)\sum_{v\in \mathcal{V}}1/\sqrt{|\mathcal{N}(v)|}$.
\end{theorem}

The proof is provided in the supplementary material. 
The lower bounds contain two parts. The first part is derived from the lower bounds in \citet{DBLP:conf/nips/Shamir14} for a class of online algorithms. The second part handles the effect of delays by showing that the individual regret of each agent cannot be smaller than the regret of a single agent with delayed full information.

We note that the individual regret of Algorithm~\ref{algo:CFTRL} is optimal with respect to scaling with the number of arms $K$ and the average regret of Algorithm~\ref{algo:DFTRL} is optimal with respect to scaling with delay $d$.

\section{Numerical experiments}
\label{sec::numerical-experiements}

In this section, we present results of numerical experiments whose goal is to compare performance of CFTRL and DFTRL algorithms with some state-of-the-art algorithms and demonstrate the tightness of our theoretical bounds. We consider the classic stochastic multi-armed bandit problem with agents communicating via different networks.

The stochastic multi-armed bandit problem is defined as follows: each arm $i$ is associated with a Bernoulli distribution with mean $\mu_i$ for $i=1,2,\dots, K$. The loss $\ell_t(i)$ from choosing arm $i$ at time step $t$ is sampled independently from the corresponding Bernoulli distribution.
In our experiments, we set $\mu_i = (1+8(i-1)/(K-1))/10$ so that $\mu_1, \dots, \mu_K$ is a linearly decreasing sequence.
Each problem instance is specified by a tuple $(K, G, d)$. The two baseline algorithms we choose are the center-based Exp3 algorithm in \citet{DBLP:conf/nips/Bar-OnM19} and the Exp3-Coop algorithm in \citet{DBLP:journals/jmlr/Cesa-BianchiGM19} whose regret upper bounds are suboptimal as discussed in the introduction.

The numerical results are for four experiments whose goals are as follows:
\begin{itemize}
    \item the first experiment compares the performance of CFTRL, DFTRL and the baselines when the number of arms increases,
    \item the second experiment validates the effect of the graph degree in the regret upper bound on CFTRL,
    \item the third experiment validates the effect of the delay parameter on the regret upper bounds of CFTRL and DFTRL, and, finally,
    \item the fourth experiment compares CFTRL and the center-based Exp3 algorithm on some sparse random graphs.
\end{itemize}

In summary, our numerical results validate theoretical results and demonstrate that CFTRL and DFTRL can achieve significant performance gains over some previously proposed algorithms.
The code for producing our experimental results is available online in the GitHub repository: \href{https://anonymous.4open.science/r/On-Regret-Optimal-Cooperative-Nonstochastic-Multi-armed-Bandits-4C76/}{[link]}.

\begin{figure}[t]
    \centering
    \includegraphics[width=.4\textwidth]{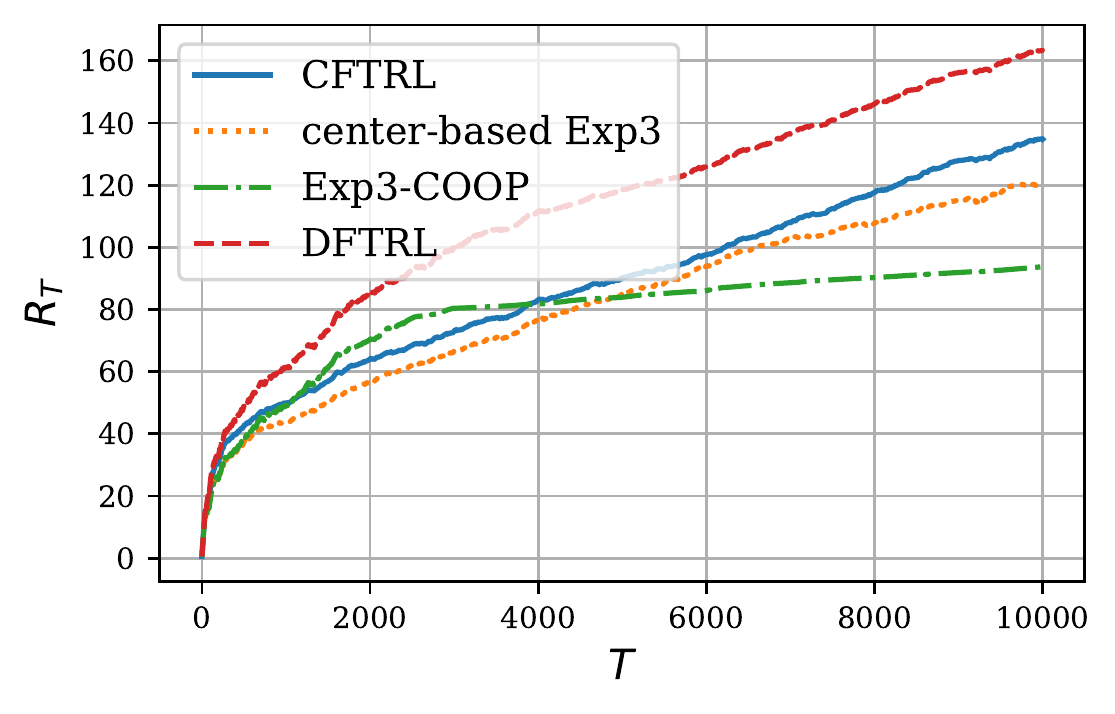}
    \includegraphics[width=.4\textwidth]{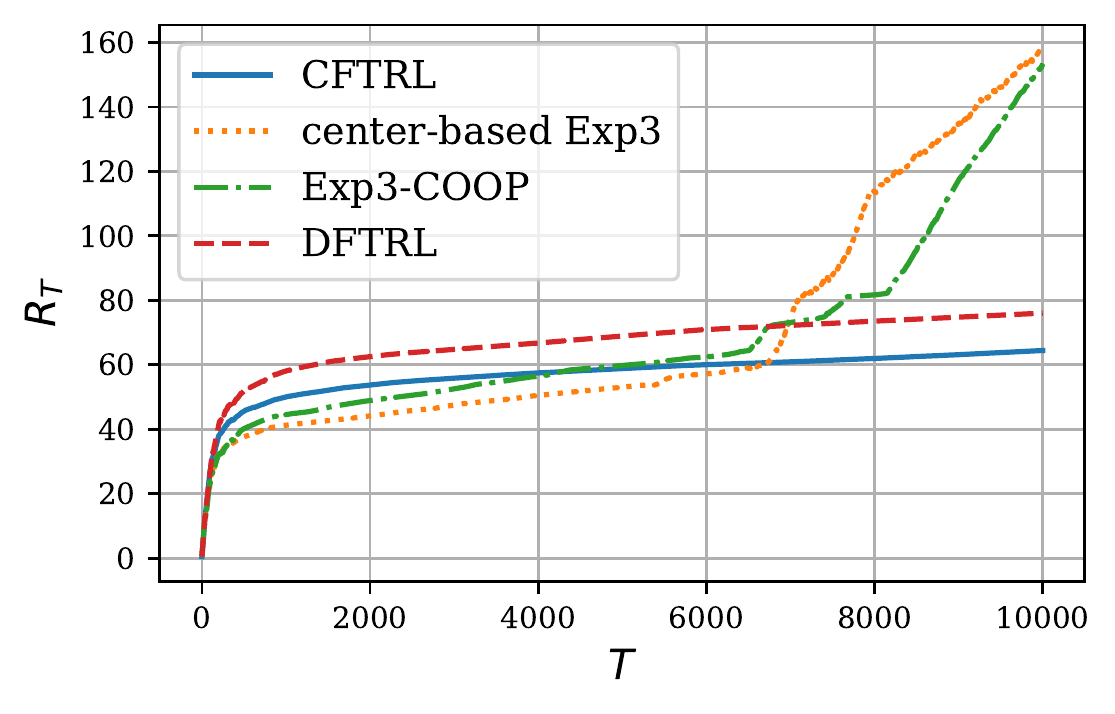}
    \includegraphics[width=.4\textwidth]{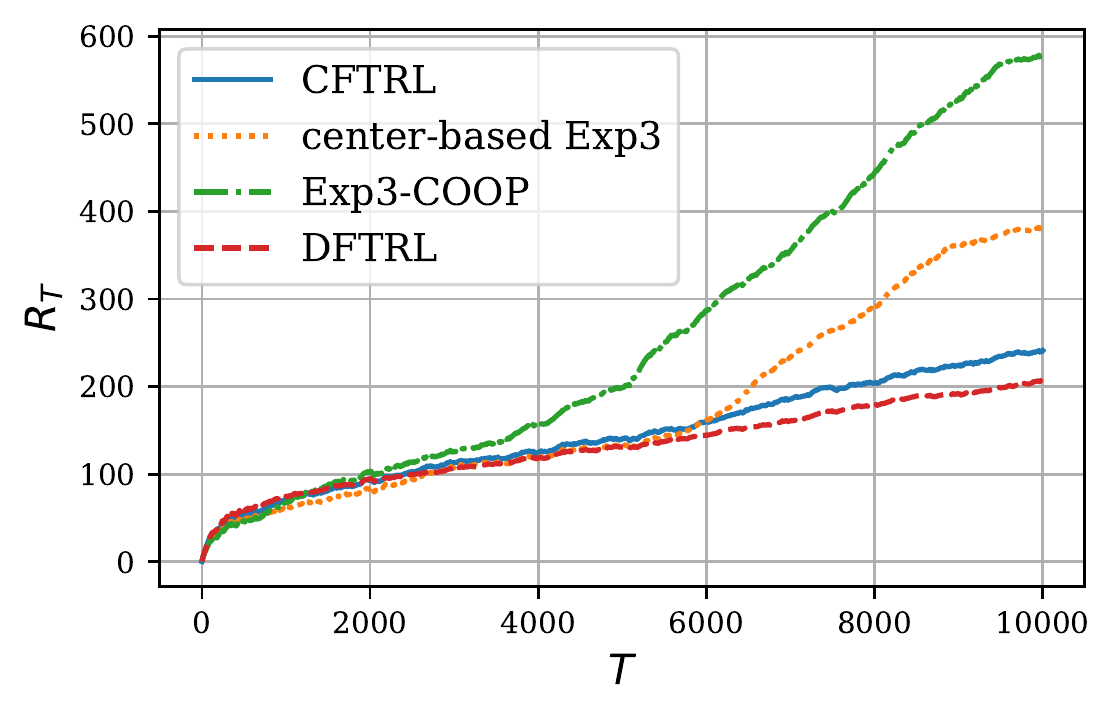}
    \caption{Average regret $R_T$ versus $T$ for different algorithms on a $2$-regular graph with $N=3$ agents and edge-delay $d=1$, and varied number of arms: (top) $K=20$ (middle) $K=30$, and (top) $K=40$. We used 10 independent simulation runs.
    }
    \label{fig:regret}
\end{figure}

\subsection{The effect of the number of arms}

In the first experiment, we evaluate the performance of CFTRL and DFTRL against the baselines on a $r$-regular graph (all nodes have the same degree of $r$). Note that in a regular graph, each agent has equal probability to be a center agent. For a $r$-regular graph, CFTRL has an individual regret upper bound of $O(\sqrt{(1/r)}\sqrt{KT})$ and DFTRL has an average regret upper bound of $O(\sqrt[4]{1-r/N}\sqrt{KT})$ when the number of arms $K$ is large enough according to our analysis. 

We first demonstrate numerical results showing that CFTRL and DFTRL can achieve significant performance gains over the center-based Exp3 algorithm whose individual regret upper bound is $O(\sqrt{(1/r)\log(K)}\sqrt{KT})$ and the Exp3-Coop algorithm whose average regret upper bound is $O(\sqrt{(1-r/N)\log(K)}\sqrt{KT})$ when the number of arms $K$ is large enough.

Figure~\ref{fig:regret} shows the regret $R_T$ versus $T$ for different number of arms, namely $20$, $30$ and $40$. The results demonstrate that CFTRL and DFTRL achieve better regret than Exp3-COOP and center-based Exp3 when the number of arms is large enough.

\subsection{The effect of graph degree on CFTRL}

\begin{figure}[t!]
    \centering
    \includegraphics[width=.4\textwidth]{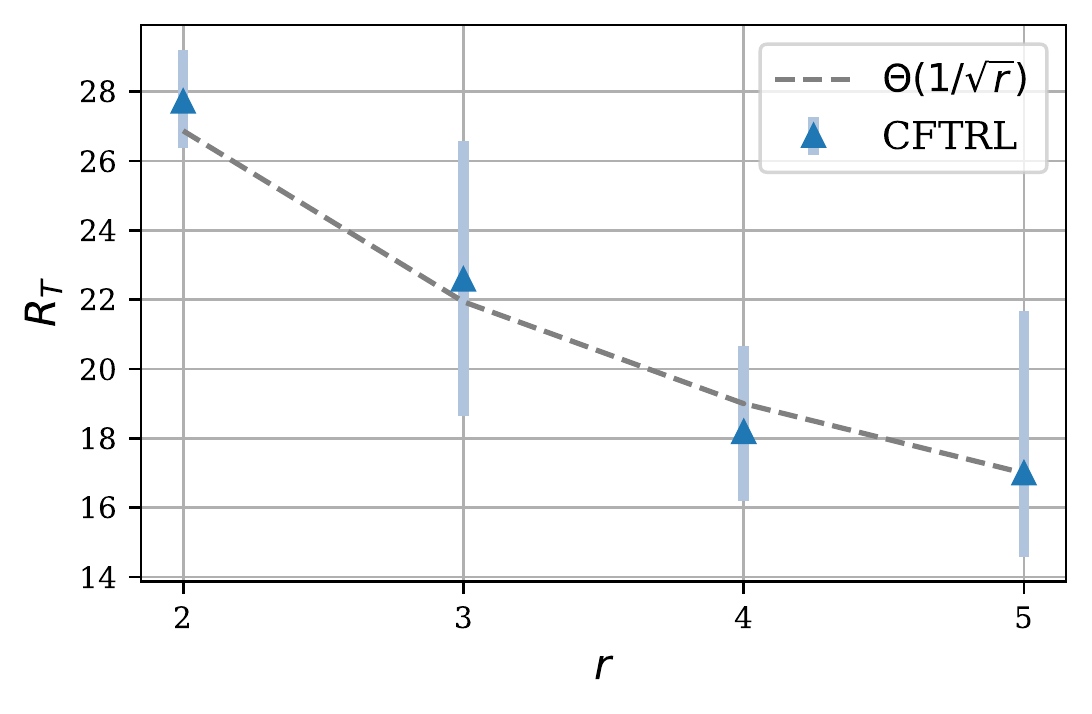}
    \caption{Average regret of CFTRL versus graph degree $r$, for a $r$-regular graph with $N=6$ nodes, $K = 10$ and $d = 1$.}
    \label{fig:degree}
\end{figure}

In the second experiment, we validate the scaling of the graph degree in the regret upper bound of CFTRL.
On the $r$-regular graph, CFTRL has a regret that scales as $O(1/\sqrt{r})$ according to Theorem~\ref{thm::CFTRL}.
We run CFTRL on the problem instances with fixed number of arms $K$, delay $d$ and increasing node degree $r$.
The results in Figure~\ref{fig:degree} shows that the averaged regret decreases as the graph degree increases and the rate of decrease is approximately $O(1/\sqrt{r})$.

\subsection{The effect of delay on CFTRL and DFTRL}

\begin{figure}[t!]
    \centering
    \includegraphics[width=.4\textwidth]{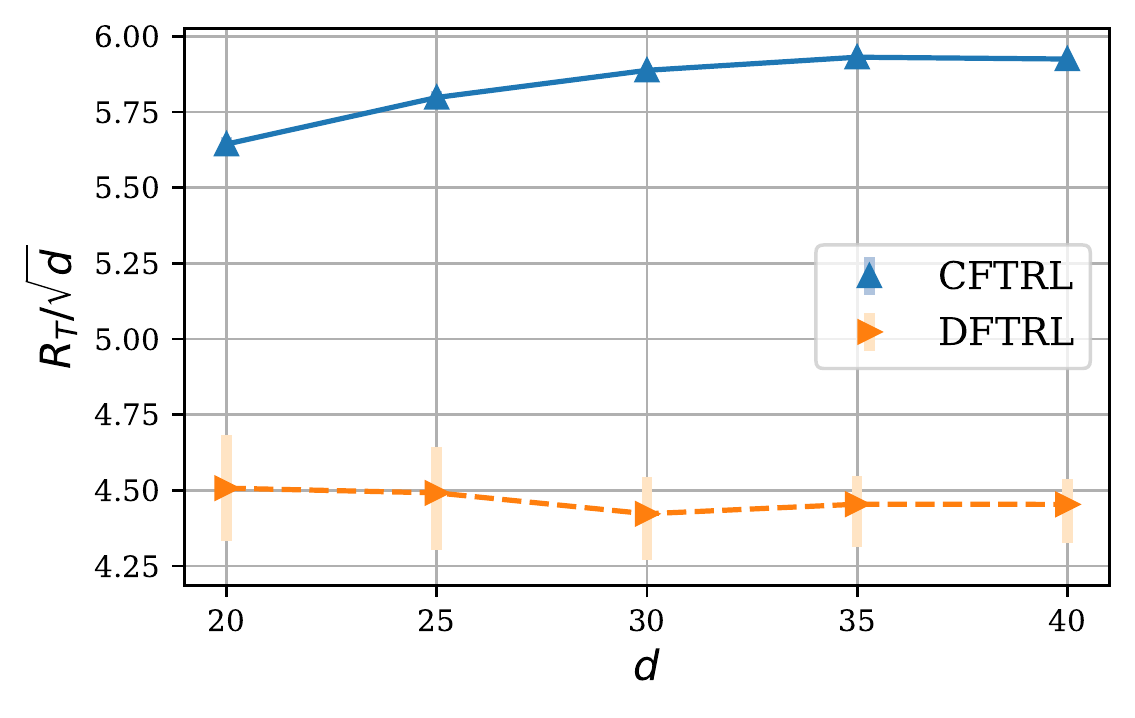}
    \caption{Average regret of CFTRL and DFTRL versus the edge-delay $d$ on a star regular graph with $N = 20$ and $K = 3$.
    }
    \label{fig:delay}
\end{figure}

In the third experiment, we run CFTRL and DFTRL algorithms on a fixed star graph $G$ with a fixed number of arms $K$ and varied edge-delay $d$ . Figure~\ref{fig:delay} shows that the normalized regret of CFTRL is $R_T / \sqrt{d} = O(\sqrt{d})$ while the normalized regret of DFTRL is $R_T / \sqrt{d} = O(1)$. Hence, when the delay $d$ is large enough, CFTRL has a linearly increasing regret with respect to $d$ which is in contrast to the sub-linear increasing regret of DFTRL.
This is consistent with our theoretical analysis, which states that CFTRL has a regret upper bound of $O(d)$ and DFTRL has a regret upper bound of $O(\sqrt{d})$. 

\subsection{The effect of graph sparsity}

\begin{figure}
    \centering
    \includegraphics[width=.4\textwidth]{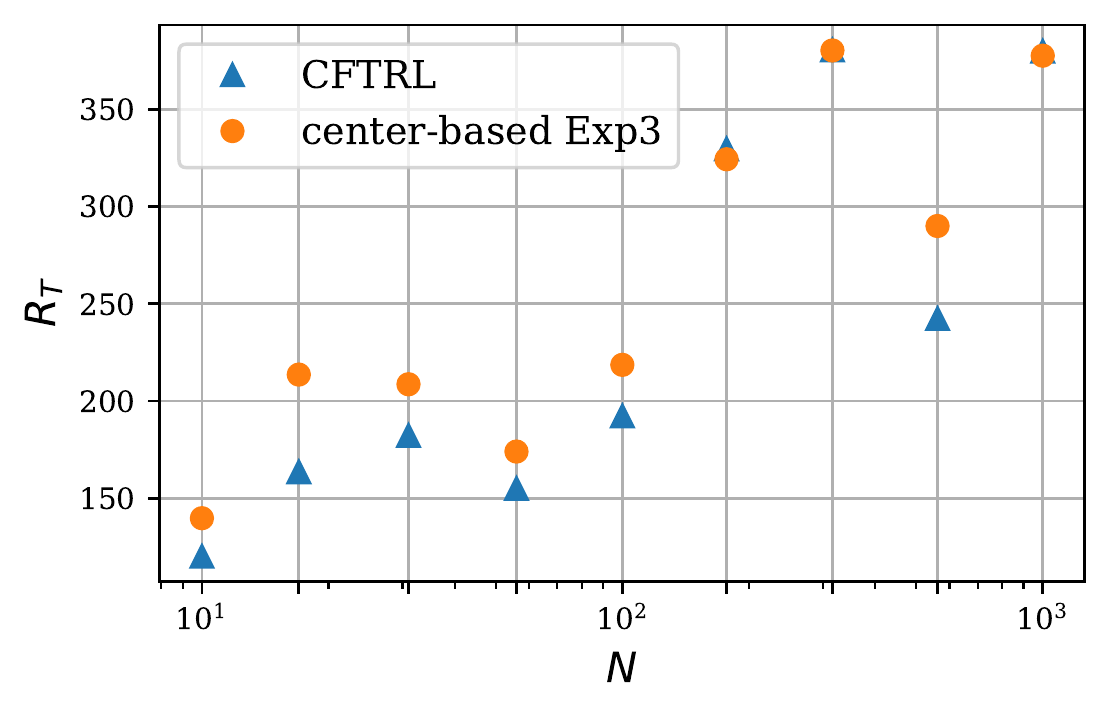}
    \caption{Average regret $R_T$ versus the number of nodes $N$ for sparse Erdős–Rényi random graphs, for CFTRL and center-based Exp3 algorithms.}
    \label{fig:erdos}
\end{figure}

In the fourth experiment, we validate that our CFTRL algorithm can outperform the center-based Exp3 algorithm on some random graphs.
We consider Erdős–Rényi random graphs of $N$ nodes with probability of an edge equal to $2\log(N)/N$. This condition ensures that the graph is connected 
and 
$|\mathcal{N}(v)| = O(\log(N)))$ for all $v\in \mathcal{V}$, almost surely \citep[Corollary 8.2]{blum_hopcroft_kannan_2020}. 
This random graph allows us to evaluate performance of algorithms for a large sparse random graph. We fix $K$ and $d$ and vary the number of nodes $N$ and compare the performance of CFTRL and the center-based Exp3 algorithm on these graphs. 
We fix $T$ to $1000$ time steps.
According to our analysis, CFTRL has a lower individual regret bound than the center-based Exp3 algorithm 
when the number of arms is large enough relative to the number of agents.  
The results in Figure~\ref{fig:erdos} indicate that CFTRL has at least as good performance as the center-based Exp3 algorithm when $K$ varies, and can have significantly better performance when the number of arms is large relative to the number of agents.

\section{Conclusion}
\label{sec::conclusion}

We presented new results for the collaborative multi-agent nonstochastic multi-armed bandit with communication delays.
We showed a lower bound on the regret of each individual agent and proposed two algorithms (CFTRL and DFTRL) together with their regret upper bounds. CFTRL provides an optimal regret of each individual agent with respect to the scaling with the number of arms. 
DFTRL has an optimal average regret with respect to the scaling with the edge-delay. Our numerical results validate our theoretical bounds and demonstrate that significant performance gains can be achieved by our two algorithms compared to state-of-the-art algorithms. 

There are several open research questions for future research. The first question is to consider the existence of a decentralized algorithm which can provide  $O((1/\sqrt{|\mathcal{N}(v)|})\sqrt{KT})$ individual regret for each agent $v$. It is unclear whether a center-based communication protocol is necessary to achieve this regret. The second question is to consider whether an algorithm exists with an optimal scaling with the number of arms and the edge-delay parameter. The third question is to understand the effect of edge-delay heterogeneity on individual regrets of agents. 

\bibliographystyle{ACM-Reference-Format}
\bibliography{references}


\begin{thebibliography}{23}


\ifx \showCODEN    \undefined \def \showCODEN     #1{\unskip}     \fi
\ifx \showDOI      \undefined \def \showDOI       #1{#1}\fi
\ifx \showISBNx    \undefined \def \showISBNx     #1{\unskip}     \fi
\ifx \showISBNxiii \undefined \def \showISBNxiii  #1{\unskip}     \fi
\ifx \showISSN     \undefined \def \showISSN      #1{\unskip}     \fi
\ifx \showLCCN     \undefined \def \showLCCN      #1{\unskip}     \fi
\ifx \shownote     \undefined \def \shownote      #1{#1}          \fi
\ifx \showarticletitle \undefined \def \showarticletitle #1{#1}   \fi
\ifx \showURL      \undefined \def \showURL       {\relax}        \fi
\providecommand\bibfield[2]{#2}
\providecommand\bibinfo[2]{#2}
\providecommand\natexlab[1]{#1}
\providecommand\showeprint[2][]{arXiv:#2}

\bibitem[Awerbuch and Kleinberg(2008)]%
        {DBLP:journals/jcss/AwerbuchK08a}
\bibfield{author}{\bibinfo{person}{Baruch Awerbuch} {and}
  \bibinfo{person}{Robert Kleinberg}.} \bibinfo{year}{2008}\natexlab{}.
\newblock \showarticletitle{Competitive collaborative learning}.
\newblock \bibinfo{journal}{\emph{J. Comput. Syst. Sci.}} \bibinfo{volume}{74},
  \bibinfo{number}{8} (\bibinfo{year}{2008}), \bibinfo{pages}{1271--1288}.
\newblock


\bibitem[Bar{-}On and Mansour(2019)]%
        {DBLP:conf/nips/Bar-OnM19}
\bibfield{author}{\bibinfo{person}{Yogev Bar{-}On} {and}
  \bibinfo{person}{Yishay Mansour}.} \bibinfo{year}{2019}\natexlab{}.
\newblock \showarticletitle{Individual Regret in Cooperative Nonstochastic
  Multi-Armed Bandits}. In \bibinfo{booktitle}{\emph{Advances in Neural
  Information Processing Systems 32: Annual Conference on Neural Information
  Processing Systems 2019, NeurIPS 2019, December 8-14, 2019, Vancouver, BC,
  Canada}}, \bibfield{editor}{\bibinfo{person}{Hanna~M. Wallach},
  \bibinfo{person}{Hugo Larochelle}, \bibinfo{person}{Alina Beygelzimer},
  \bibinfo{person}{Florence d'Alch{\'{e}}{-}Buc}, \bibinfo{person}{Emily~B.
  Fox}, {and} \bibinfo{person}{Roman Garnett}} (Eds.).
  \bibinfo{pages}{3110--3120}.
\newblock


\bibitem[Baraglia et~al\mbox{.}(2013)]%
        {DBLP:journals/jcss/BaragliaDMR13}
\bibfield{author}{\bibinfo{person}{Ranieri Baraglia}, \bibinfo{person}{Patrizio
  Dazzi}, \bibinfo{person}{Matteo Mordacchini}, {and} \bibinfo{person}{Laura
  Ricci}.} \bibinfo{year}{2013}\natexlab{}.
\newblock \showarticletitle{A peer-to-peer recommender system for self-emerging
  user communities based on gossip overlays}.
\newblock \bibinfo{journal}{\emph{J. Comput. Syst. Sci.}} \bibinfo{volume}{79},
  \bibinfo{number}{2} (\bibinfo{year}{2013}), \bibinfo{pages}{291--308}.
\newblock


\bibitem[Blum et~al\mbox{.}(2020)]%
        {blum_hopcroft_kannan_2020}
\bibfield{author}{\bibinfo{person}{Avrim Blum}, \bibinfo{person}{John
  Hopcroft}, {and} \bibinfo{person}{Ravindran Kannan}.}
  \bibinfo{year}{2020}\natexlab{}.
\newblock \bibinfo{booktitle}{\emph{Foundations of Data Science}}.
\newblock \bibinfo{publisher}{Cambridge University Press}.
\newblock


\bibitem[Cesa{-}Bianchi et~al\mbox{.}(2020)]%
        {DBLP:conf/alt/Cesa-BianchiCM20}
\bibfield{author}{\bibinfo{person}{Nicol{\`{o}} Cesa{-}Bianchi},
  \bibinfo{person}{Tommaso Cesari}, {and} \bibinfo{person}{Claire Monteleoni}.}
  \bibinfo{year}{2020}\natexlab{}.
\newblock \showarticletitle{Cooperative Online Learning: Keeping your Neighbors
  Updated}. In \bibinfo{booktitle}{\emph{Algorithmic Learning Theory, {ALT}
  2020, 8-11 February 2020, San Diego, CA, {USA}}}
  \emph{(\bibinfo{series}{Proceedings of Machine Learning Research},
  Vol.~\bibinfo{volume}{117})}, \bibfield{editor}{\bibinfo{person}{Aryeh
  Kontorovich} {and} \bibinfo{person}{Gergely Neu}} (Eds.).
  \bibinfo{publisher}{{PMLR}}, \bibinfo{pages}{234--250}.
\newblock


\bibitem[Cesa{-}Bianchi et~al\mbox{.}(2019)]%
        {DBLP:journals/jmlr/Cesa-BianchiGM19}
\bibfield{author}{\bibinfo{person}{Nicol{\`{o}} Cesa{-}Bianchi},
  \bibinfo{person}{Claudio Gentile}, {and} \bibinfo{person}{Yishay Mansour}.}
  \bibinfo{year}{2019}\natexlab{}.
\newblock \showarticletitle{Delay and Cooperation in Nonstochastic Bandits}.
\newblock \bibinfo{journal}{\emph{J. Mach. Learn. Res.}}  \bibinfo{volume}{20}
  (\bibinfo{year}{2019}), \bibinfo{pages}{17:1--17:38}.
\newblock


\bibitem[Chakravarty et~al\mbox{.}(2014)]%
        {chakravarty_mitra_sarkar_2014}
\bibfield{author}{\bibinfo{person}{Satya~R. Chakravarty},
  \bibinfo{person}{Manipushpak Mitra}, {and} \bibinfo{person}{Palash Sarkar}.}
  \bibinfo{year}{2014}\natexlab{}.
\newblock \bibinfo{booktitle}{\emph{A Course on Cooperative Game Theory}}.
\newblock \bibinfo{publisher}{Cambridge University Press}.
\newblock


\bibitem[Flaspohler et~al\mbox{.}(2021)]%
        {DBLP:conf/icml/FlaspohlerOCMOO21}
\bibfield{author}{\bibinfo{person}{Genevieve Flaspohler},
  \bibinfo{person}{Francesco Orabona}, \bibinfo{person}{Judah Cohen},
  \bibinfo{person}{Soukayna Mouatadid}, \bibinfo{person}{Miruna Oprescu},
  \bibinfo{person}{Paulo Orenstein}, {and} \bibinfo{person}{Lester Mackey}.}
  \bibinfo{year}{2021}\natexlab{}.
\newblock \showarticletitle{Online Learning with Optimism and Delay}. In
  \bibinfo{booktitle}{\emph{Proceedings of the 38th International Conference on
  Machine Learning, {ICML} 2021, 18-24 July 2021, Virtual Event}}
  \emph{(\bibinfo{series}{Proceedings of Machine Learning Research},
  Vol.~\bibinfo{volume}{139})}, \bibfield{editor}{\bibinfo{person}{Marina
  Meila} {and} \bibinfo{person}{Tong Zhang}} (Eds.).
  \bibinfo{publisher}{{PMLR}}, \bibinfo{pages}{3363--3373}.
\newblock


\bibitem[Hagberg et~al\mbox{.}(2008)]%
        {hagberg2008exploring}
\bibfield{author}{\bibinfo{person}{Aric Hagberg}, \bibinfo{person}{Pieter
  Swart}, {and} \bibinfo{person}{Daniel S~Chult}.}
  \bibinfo{year}{2008}\natexlab{}.
\newblock \bibinfo{booktitle}{\emph{Exploring network structure, dynamics, and
  function using NetworkX}}.
\newblock \bibinfo{type}{{T}echnical {R}eport}. \bibinfo{institution}{Los
  Alamos National Lab.(LANL), Los Alamos, NM (United States)}.
\newblock


\bibitem[Harris et~al\mbox{.}(2020)]%
        {harris2020array}
\bibfield{author}{\bibinfo{person}{Charles~R. Harris},
  \bibinfo{person}{K.~Jarrod Millman}, \bibinfo{person}{St{\'{e}}fan~J. van~der
  Walt}, \bibinfo{person}{Ralf Gommers}, \bibinfo{person}{Pauli Virtanen},
  \bibinfo{person}{David Cournapeau}, \bibinfo{person}{Eric Wieser},
  \bibinfo{person}{Julian Taylor}, \bibinfo{person}{Sebastian Berg},
  \bibinfo{person}{Nathaniel~J. Smith}, \bibinfo{person}{Robert Kern},
  \bibinfo{person}{Matti Picus}, \bibinfo{person}{Stephan Hoyer},
  \bibinfo{person}{Marten~H. van Kerkwijk}, \bibinfo{person}{Matthew Brett},
  \bibinfo{person}{Allan Haldane}, \bibinfo{person}{Jaime~Fern{\'{a}}ndez del
  R{\'{i}}o}, \bibinfo{person}{Mark Wiebe}, \bibinfo{person}{Pearu Peterson},
  \bibinfo{person}{Pierre G{\'{e}}rard-Marchant}, \bibinfo{person}{Kevin
  Sheppard}, \bibinfo{person}{Tyler Reddy}, \bibinfo{person}{Warren Weckesser},
  \bibinfo{person}{Hameer Abbasi}, \bibinfo{person}{Christoph Gohlke}, {and}
  \bibinfo{person}{Travis~E. Oliphant}.} \bibinfo{year}{2020}\natexlab{}.
\newblock \showarticletitle{Array programming with {NumPy}}.
\newblock \bibinfo{journal}{\emph{Nature}} \bibinfo{volume}{585},
  \bibinfo{number}{7825} (\bibinfo{date}{Sept.} \bibinfo{year}{2020}),
  \bibinfo{pages}{357--362}.
\newblock


\bibitem[Joulani et~al\mbox{.}(2013)]%
        {DBLP:conf/icml/JoulaniGS13}
\bibfield{author}{\bibinfo{person}{Pooria Joulani},
  \bibinfo{person}{Andr{\'{a}}s Gy{\"{o}}rgy}, {and} \bibinfo{person}{Csaba
  Szepesv{\'{a}}ri}.} \bibinfo{year}{2013}\natexlab{}.
\newblock \showarticletitle{Online Learning under Delayed Feedback}. In
  \bibinfo{booktitle}{\emph{Proceedings of the 30th International Conference on
  Machine Learning, {ICML} 2013, Atlanta, GA, USA, 16-21 June 2013}}
  \emph{(\bibinfo{series}{{JMLR} Workshop and Conference Proceedings},
  Vol.~\bibinfo{volume}{28})}. \bibinfo{publisher}{JMLR.org},
  \bibinfo{pages}{1453--1461}.
\newblock


\bibitem[Kar et~al\mbox{.}(2011)]%
        {DBLP:conf/cdc/KarPC11}
\bibfield{author}{\bibinfo{person}{Soummya Kar}, \bibinfo{person}{H.~Vincent
  Poor}, {and} \bibinfo{person}{Shuguang Cui}.}
  \bibinfo{year}{2011}\natexlab{}.
\newblock \showarticletitle{Bandit problems in networks: Asymptotically
  efficient distributed allocation rules}. In \bibinfo{booktitle}{\emph{50th
  {IEEE} Conference on Decision and Control and European Control Conference,
  11th European Control Conference, {CDC/ECC} 2011, Orlando, FL, USA, December
  12-15, 2011}}. \bibinfo{publisher}{{IEEE}}, \bibinfo{pages}{1771--1778}.
\newblock


\bibitem[Kolla et~al\mbox{.}(2018)]%
        {DBLP:journals/ton/KollaJG18}
\bibfield{author}{\bibinfo{person}{Ravi~Kumar Kolla},
  \bibinfo{person}{Krishna~P. Jagannathan}, {and} \bibinfo{person}{Aditya
  Gopalan}.} \bibinfo{year}{2018}\natexlab{}.
\newblock \showarticletitle{Collaborative Learning of Stochastic Bandits Over a
  Social Network}.
\newblock \bibinfo{journal}{\emph{{IEEE/ACM} Trans. Netw.}}
  \bibinfo{volume}{26}, \bibinfo{number}{4} (\bibinfo{year}{2018}),
  \bibinfo{pages}{1782--1795}.
\newblock


\bibitem[Lanctot et~al\mbox{.}(2017)]%
        {DBLP:conf/nips/LanctotZGLTPSG17}
\bibfield{author}{\bibinfo{person}{Marc Lanctot},
  \bibinfo{person}{Vin{\'{\i}}cius~Flores Zambaldi}, \bibinfo{person}{Audrunas
  Gruslys}, \bibinfo{person}{Angeliki Lazaridou}, \bibinfo{person}{Karl Tuyls},
  \bibinfo{person}{Julien P{\'{e}}rolat}, \bibinfo{person}{David Silver}, {and}
  \bibinfo{person}{Thore Graepel}.} \bibinfo{year}{2017}\natexlab{}.
\newblock \showarticletitle{A Unified Game-Theoretic Approach to Multiagent
  Reinforcement Learning}. In \bibinfo{booktitle}{\emph{Advances in Neural
  Information Processing Systems 30: Annual Conference on Neural Information
  Processing Systems 2017, December 4-9, 2017, Long Beach, CA, {USA}}},
  \bibfield{editor}{\bibinfo{person}{Isabelle Guyon}, \bibinfo{person}{Ulrike
  von Luxburg}, \bibinfo{person}{Samy Bengio}, \bibinfo{person}{Hanna~M.
  Wallach}, \bibinfo{person}{Rob Fergus}, \bibinfo{person}{S.~V.~N.
  Vishwanathan}, {and} \bibinfo{person}{Roman Garnett}} (Eds.).
  \bibinfo{pages}{4190--4203}.
\newblock


\bibitem[Landgren et~al\mbox{.}(2016)]%
        {DBLP:conf/cdc/LandgrenSL16}
\bibfield{author}{\bibinfo{person}{Peter Landgren}, \bibinfo{person}{Vaibhav
  Srivastava}, {and} \bibinfo{person}{Naomi~Ehrich Leonard}.}
  \bibinfo{year}{2016}\natexlab{}.
\newblock \showarticletitle{Distributed cooperative decision-making in
  multiarmed bandits: Frequentist and Bayesian algorithms}. In
  \bibinfo{booktitle}{\emph{55th {IEEE} Conference on Decision and Control,
  {CDC} 2016, Las Vegas, NV, USA, December 12-14, 2016}}.
  \bibinfo{publisher}{{IEEE}}, \bibinfo{pages}{167--172}.
\newblock


\bibitem[Lynch(1996)]%
        {DBLP:books/mk/Lynch96}
\bibfield{author}{\bibinfo{person}{Nancy~A. Lynch}.}
  \bibinfo{year}{1996}\natexlab{}.
\newblock \bibinfo{booktitle}{\emph{Distributed Algorithms}}.
\newblock \bibinfo{publisher}{Morgan Kaufmann}.
\newblock
\showISBNx{1-55860-348-4}


\bibitem[Mart{\'{\i}}nez{-}Rubio et~al\mbox{.}(2019)]%
        {DBLP:conf/nips/Martinez-RubioK19}
\bibfield{author}{\bibinfo{person}{David Mart{\'{\i}}nez{-}Rubio},
  \bibinfo{person}{Varun Kanade}, {and} \bibinfo{person}{Patrick Rebeschini}.}
  \bibinfo{year}{2019}\natexlab{}.
\newblock \showarticletitle{Decentralized Cooperative Stochastic Bandits}. In
  \bibinfo{booktitle}{\emph{Advances in Neural Information Processing Systems
  32: Annual Conference on Neural Information Processing Systems 2019, NeurIPS
  2019, December 8-14, 2019, Vancouver, BC, Canada}},
  \bibfield{editor}{\bibinfo{person}{Hanna~M. Wallach}, \bibinfo{person}{Hugo
  Larochelle}, \bibinfo{person}{Alina Beygelzimer}, \bibinfo{person}{Florence
  d'Alch{\'{e}}{-}Buc}, \bibinfo{person}{Emily~B. Fox}, {and}
  \bibinfo{person}{Roman Garnett}} (Eds.). \bibinfo{pages}{4531--4542}.
\newblock


\bibitem[Shamir(2014)]%
        {DBLP:conf/nips/Shamir14}
\bibfield{author}{\bibinfo{person}{Ohad Shamir}.}
  \bibinfo{year}{2014}\natexlab{}.
\newblock \showarticletitle{Fundamental Limits of Online and Distributed
  Algorithms for Statistical Learning and Estimation}. In
  \bibinfo{booktitle}{\emph{Advances in Neural Information Processing Systems
  27: Annual Conference on Neural Information Processing Systems 2014, December
  8-13 2014, Montreal, Quebec, Canada}},
  \bibfield{editor}{\bibinfo{person}{Zoubin Ghahramani}, \bibinfo{person}{Max
  Welling}, \bibinfo{person}{Corinna Cortes}, \bibinfo{person}{Neil~D.
  Lawrence}, {and} \bibinfo{person}{Kilian~Q. Weinberger}} (Eds.).
  \bibinfo{pages}{163--171}.
\newblock


\bibitem[Sz{\"{o}}r{\'{e}}nyi et~al\mbox{.}(2013)]%
        {DBLP:conf/icml/SzorenyiBHOJK13}
\bibfield{author}{\bibinfo{person}{Bal{\'{a}}zs Sz{\"{o}}r{\'{e}}nyi},
  \bibinfo{person}{R{\'{o}}bert Busa{-}Fekete}, \bibinfo{person}{Istv{\'{a}}n
  Heged{\"{u}}s}, \bibinfo{person}{R{\'{o}}bert Orm{\'{a}}ndi},
  \bibinfo{person}{M{\'{a}}rk Jelasity}, {and} \bibinfo{person}{Bal{\'{a}}zs
  K{\'{e}}gl}.} \bibinfo{year}{2013}\natexlab{}.
\newblock \showarticletitle{Gossip-based distributed stochastic bandit
  algorithms}. In \bibinfo{booktitle}{\emph{Proceedings of the 30th
  International Conference on Machine Learning, {ICML} 2013, Atlanta, GA, USA,
  16-21 June 2013}} \emph{(\bibinfo{series}{{JMLR} Workshop and Conference
  Proceedings}, Vol.~\bibinfo{volume}{28})}. \bibinfo{publisher}{JMLR.org},
  \bibinfo{pages}{19--27}.
\newblock


\bibitem[Thune et~al\mbox{.}(2019)]%
        {DBLP:conf/nips/ThuneCS19}
\bibfield{author}{\bibinfo{person}{Tobias~Sommer Thune},
  \bibinfo{person}{Nicol{\`{o}} Cesa{-}Bianchi}, {and} \bibinfo{person}{Yevgeny
  Seldin}.} \bibinfo{year}{2019}\natexlab{}.
\newblock \showarticletitle{Nonstochastic Multiarmed Bandits with Unrestricted
  Delays}. In \bibinfo{booktitle}{\emph{Advances in Neural Information
  Processing Systems 32: Annual Conference on Neural Information Processing
  Systems 2019, NeurIPS 2019, December 8-14, 2019, Vancouver, BC, Canada}},
  \bibfield{editor}{\bibinfo{person}{Hanna~M. Wallach}, \bibinfo{person}{Hugo
  Larochelle}, \bibinfo{person}{Alina Beygelzimer}, \bibinfo{person}{Florence
  d'Alch{\'{e}}{-}Buc}, \bibinfo{person}{Emily~B. Fox}, {and}
  \bibinfo{person}{Roman Garnett}} (Eds.). \bibinfo{pages}{6538--6547}.
\newblock


\bibitem[Tsallis(1988)]%
        {tsallis1988possible}
\bibfield{author}{\bibinfo{person}{Constantino Tsallis}.}
  \bibinfo{year}{1988}\natexlab{}.
\newblock \showarticletitle{Possible generalization of Boltzmann-Gibbs
  statistics}.
\newblock \bibinfo{journal}{\emph{Journal of statistical physics}}
  \bibinfo{volume}{52}, \bibinfo{number}{1} (\bibinfo{year}{1988}),
  \bibinfo{pages}{479--487}.
\newblock


\bibitem[Zimmert and Seldin(2020)]%
        {DBLP:conf/aistats/ZimmertS20}
\bibfield{author}{\bibinfo{person}{Julian Zimmert} {and}
  \bibinfo{person}{Yevgeny Seldin}.} \bibinfo{year}{2020}\natexlab{}.
\newblock \showarticletitle{An Optimal Algorithm for Adversarial Bandits with
  Arbitrary Delays}. In \bibinfo{booktitle}{\emph{The 23rd International
  Conference on Artificial Intelligence and Statistics, {AISTATS} 2020, 26-28
  August 2020, Online [Palermo, Sicily, Italy]}}
  \emph{(\bibinfo{series}{Proceedings of Machine Learning Research},
  Vol.~\bibinfo{volume}{108})}, \bibfield{editor}{\bibinfo{person}{Silvia
  Chiappa} {and} \bibinfo{person}{Roberto Calandra}} (Eds.).
  \bibinfo{publisher}{{PMLR}}, \bibinfo{pages}{3285--3294}.
\newblock


\bibitem[Zimmert and Seldin(2021)]%
        {DBLP:journals/jmlr/ZimmertS21}
\bibfield{author}{\bibinfo{person}{Julian Zimmert} {and}
  \bibinfo{person}{Yevgeny Seldin}.} \bibinfo{year}{2021}\natexlab{}.
\newblock \showarticletitle{Tsallis-INF: An Optimal Algorithm for Stochastic
  and Adversarial Bandits}.
\newblock \bibinfo{journal}{\emph{J. Mach. Learn. Res.}}  \bibinfo{volume}{22}
  (\bibinfo{year}{2021}), \bibinfo{pages}{28:1--28:49}.
\newblock


\end{thebibliography}


\section*{Appendix}

\subsection{Upper bounds}

\subsubsection{An auxiliary lemma}

We show and prove the next lemma which is used for proving some other lemmas.

\begin{lemma}
\label{lm:conditional_exp}
For all $1\leq s \leq t$, $v\in \mathcal{V}$ and $i\in \mathcal{A}$, it holds
\begin{enumerate}
    \item $\mathbb{E}_{s}[\hat{\ell}_t^v(i)] = \ell_t(i)$
    \item $\mathbb{E}_t[\hat{\ell}_t^v(i)^2] = \ell_t(i)^2 / q_t^v(i)$, and 
    \item $\mathbb{E}_t[\hat{\ell}_t^v(i) \hat{L}_{t}^{v,miss}(i)] = \ell_t(i) \hat{L}_{t}^{v,miss}(i)$.
\end{enumerate}
\end{lemma}

\begin{proof}
Consider an arbitrary $v\in \mathcal{V}$. By definition, $p_t^v$ is determined by $\hat{L}_{t}^{v, obs}$ which depends on the realization of $I_1(v), I_2(v),\dots,I_{t-d-1}(v)$. Note that the following equations hold, for all $i\in \mathcal{A}$,
\begin{equation*}
\begin{aligned}
    \mathbb{E}_{t}[\hat{\ell}_t^v(i)] &= \frac{\ell_t(i)}{q_t^v(i)}\mathbb{E}_{t}\left[ \mathbb{I}\left\{i\in\mathcal{J}_t(v)\right\}\right] = \frac{\ell_t(i)}{q_t^v(i)}q_t^v(i) = \ell_t(i) \\
    \mathbb{E}_{t}[\hat{\ell}_t^v(i)^2] &= \frac{\ell_t(i)^2}{q_t^v(i)^2}\mathbb{E}_{t}\left[ \mathbb{I}\left\{i\in\mathcal{J}_t(v)\right\}\right] = \frac{\ell_t(i)^2}{q_t^v(i)^2}q_t^v(i) = \frac{\ell_t(i)^2}{q_t^v(i)}.  
\end{aligned}
\end{equation*}
By the tower law of conditional expectation, we have  $\mathbb{E}_{s}[\hat{\ell}_t^v(i)] = \ell_t(i)$ for all $s\leq t$. Note that $\hat{L}^{v,obs}_t(i) = \sum_{s=1}^{t-d-1}\hat{\ell}^{v, obs}_s(i)$, thus, $\hat{L}_{t}^{v,miss}(i) = \sum_{s=t-d}^{t-1}\hat{\ell}_s^v(i)$, which depends on the realization of $I_1(v), I_2(v),\dots,I_{t-1}(v)$ for all $v\in \mathcal{V}$. Hence, for all $i\in \mathcal{A}$, we have
\begin{equation*}
    \mathbb{E}_t[\hat{\ell}_t^v(i) \hat{L}_{t}^{v,miss}(i)] =  \mathbb{E}_t[\hat{\ell}_t^v(i)]\hat{L}_{t}^{v,miss}(i) = \ell_t(i)\hat{L}_{t}^{v,miss}(i).
\end{equation*}
\end{proof}

\subsubsection{Proof of Lemma~3.2}
\label{sec::upper-bound-lemma-3}

Let $i^\ast = \arg\min_{i\in\mathcal{A}} \sum_{t=1}^T \ell_t(i)$ be the best action in hindsight. Consider an arbitrary agent $v\in \mathcal{V}$.
Following arguments in Theorem~3 from \cite{DBLP:conf/aistats/ZimmertS20}, we have

\begin{equation*}
\begin{aligned}
R_{T}^v
&=\mathbb{E}\left[\sum_{t=1}^{T} \ell_{t}(I_t(v))-\sum_{t=1}^{T}\ell_{t}(i^\ast)\right] \\
&=  \mathbb{E}\left[\sum_{t=1}^T\left\langle p_t^v, \hat{\ell}_{t}(v)\right\rangle-\hat{L}_{T+1}^v(i^\ast)\right]\\
&= \mathbb{E}\left[\alpha + \beta + \gamma \right] \\
\end{aligned}
\end{equation*}
where 
\begin{equation*}
\begin{aligned}
\alpha &= \sum_{t=1}^T\left[\bar{F}_{t}^{*}\left(-\hat{L}_{t}^{v,obs}-\hat{\ell}_t^v\right)-\bar{F}_{t}^{*}\left(-\hat{L}_{t}^{v,obs}\right)+\left\langle p_t^v, \hat{\ell}_t^v\right\rangle\right] \\
\beta &=  \sum_{t=1}^T\left[\bar{F}_{t}^{*}\left(-\hat{L}_{t}^{v,obs}\right)-\bar{F}_{t}^{*}\left(-\hat{L}_{t}^{v,obs}-\hat{\ell}_t^v\right) -\bar{F}_{t}^{*}\left(-\hat{L}_t^v\right) +\bar{F}_{t}^{*}\left(-\hat{L}_{t+1}^v\right)\right]\\
\gamma &= \sum_{t=1}^T\left[\bar{F}_{t}^{*}\left(-\hat{L}_t^v\right)-\bar{F}_{t}^{*}\left(-\hat{L}_{t+1}^v\right)\right]-\hat{L}_{T+1}^v(i^{*}).
\end{aligned}
\end{equation*}

Next, we bound $\alpha, \beta$ and $\gamma$, as follows.

First, for $\alpha$ we have
$$
\mathbb{E}\left[\alpha\right] 
\leq \frac{1}{2} \mathbb{E}\left[\sum_{t=1}^T \sum_{i\in \mathcal{J}_t(v)} \frac{\hat{\ell}_t^v(i)^2}{f^{\prime\prime}_t(p_t^v(i))} \right]
\leq \frac{1}{2} \mathbb{E}\left[\sum_{t=1}^T \sum_{i\in \mathcal{A}} \frac{\hat{\ell}_t^v(i)^2}{f^{\prime\prime}_t(p_t^v(i))} \right] 
\leq \frac{1}{2} \mathbb{E}\left[\sum_{t=1}^T \sum_{i\in \mathcal{A}} \frac{1}{q_t^v(i)f^{\prime\prime}_t(p_t^v(i))}\right]
$$
where the first inequality comes from the proof in Lemma~1 in \cite{DBLP:conf/aistats/ZimmertS20} and the last inequality comes from Lemma \ref{lm:conditional_exp}.

Second, by Lemma~3 in \cite{DBLP:conf/aistats/ZimmertS20}, 
\begin{eqnarray*}
&&    \bar{F}_{t}^{*}\left(-\hat{L}_{t}^{v,obs}\right)-\bar{F}_{t}^{*}\left(-\hat{L}_{t}^{v,obs}-\hat{\ell}_t^v\right) -\bar{F}_{t}^{*}\left(-\hat{L}_t^v\right) +\bar{F}_{t}^{*}\left(-\hat{L}_{t+1}^v\right)\\
&\leq & \sum_{i\in \mathcal{J}_t(v)}
\int_{0}^{1} \hat{\ell}_t^v(i) f_{t}^{\prime \prime}\left(p_{t}^v(i)\right)^{-1} \hat{L}_{t}^{v,miss}(i) d x \\
&=& \sum_{i\in \mathcal{J}_t(v)} \hat{\ell}_t^v(i) f_{t}^{\prime \prime}\left(p_{t}^v(i)\right)^{-1} \hat{L}_{t}^{v,miss}(i).
\end{eqnarray*}
Hence, for $\beta$ we have 
$$
\mathbb{E}\left[\beta\right]  
\leq \mathbb{E}\left[ \sum_{t=1}^T \sum_{i\in \mathcal{A}} \frac{\ell_t(i)\hat{L}_{t}^{v,miss}(i)}{f^{\prime\prime}_t(p_t^v(i))}  \right] 
\leq \mathbb{E}\left[ \sum_{t=1}^T \sum_{i\in \mathcal{A}} \frac{\sum_{s=1}^{d}\hat{\ell}_{t-s}^v(i)}{f^{\prime\prime}_t(p_t^v(i))} \right]
\leq d \sum_{t=1}^T \mathbb{E}\left[ \sum_{i\in \mathcal{A}} \frac{1}{f^{\prime\prime}_t(p_t^v(i))}\right]
$$
where the first and third inequalities come from Lemma \ref{lm:conditional_exp}.

Finally, by Lemma 2 in \cite{DBLP:conf/aistats/ZimmertS20}, we have

\begin{equation*}
\begin{aligned}
\gamma &\leq \max_{x \in \mathcal{P}_{K-1}}-F_{1}(x)+\sum_{t=2}^{T} \max_{x \in \mathcal{P}_{K-1}}\left(F_{t-1}(x)-F_{t}(x)\right). 
\end{aligned}
\end{equation*}

\subsubsection{Proof of Lemma~3.3}

Consider an arbitrary agent $v\in \mathcal{V}$. The arm selection distribution $p^v_t$ is the solution of the following convex optimization problem:
$$
\begin{array}{rl}
\hbox{minimize} & \eta(v) \sum_{s=1}^{t-1} \left<x,\hat{\ell}_{t}^{v, obs}\right> + \left(\sum_{i=1}^K -2\sqrt{x_i}\right)  \\
\hbox{over}     & x\in \mathbf{R}^K\\
\hbox{subject to} & x_i \geq 0, \hbox{ for all } i\in \mathcal{A}\\
& \sum_{i=1}^K x_i = 1.
\end{array}
$$

The distribution $p_t^v$ satisfies
\begin{equation}
p_t^v(i) = \frac{1}{\left(\eta(v) \sum_{s=1}^{t-1}\hat{\ell}_{t}^{v, obs}(i) + \lambda_t^v\right)^2}, \hbox{ for } i\in \mathcal{A}
\label{equ:pt}
\end{equation}
where $\lambda_t^v\geq 0$ is the Lagrange multiplier associated with the constraint $\sum_{i=1}^K p_t^v(i) = 1$, which is the solution of the following fixed-point equation
\begin{equation}
\sum_{i=1}^K \frac{1}{\left(\eta(v)\sum_{s=1}^{t-1}\hat{\ell}_{t}^{v, obs}(i) + \lambda_t^v\right)^2} = 1.
\label{equ:lambdat}
\end{equation}

From (\ref{equ:lambdat}) it follows that $\lambda_t^v$ is a decreasing sequence in $t$. To see this, note
$$
\sum_{i=1}^K \frac{1}{\left(\eta(v)\sum_{s=1}^{t-1}\hat{\ell}_{t}^{v, obs}(i) + \lambda_t^v + (\eta(v)\hat{\ell}_t^{v,obs}(i)+\lambda_{t+1}^v-\lambda_t^v)\right)^2} = 1
$$
If $\lambda_{t+1}^v - \lambda_t^v > 0$, then the left-hand side in the last equation is less than $1$ and, hence, the equation cannot hold. Since $\hat{\ell}_t^{v,obs}(i) > 0$ if, and only if, $I_t(v) = i$, it follows that $\hat{\ell}_t^{v,obs}(I_t(v)) + \lambda_{t+1}^v-\lambda_t^v \geq 0$ and $\lambda_{t+1}^v\leq \lambda_t^v$. 

From (\ref{equ:pt}), for all $i\in \mathcal{A}$,
$$
p_{t+1}^v(i) = \frac{p_t^v(i)}{\left(1+\sqrt{p_t^v(i)}(\eta(v)\hat{\ell}_t^{v,obs}(i)+\lambda_{t+1}^v-\lambda_t^v)\right)^2}.
$$
From this and the fact $\hat{\ell}_t^{v,obs}(I_t(v)) + \lambda_{t+1}^v-\lambda_t^v \geq 0$, it follows 
$$
p_{t+1}^v(I_t(v))\leq p_t^v(I_t(v)).
$$

Let $\gamma = \lambda_{t+1}^v-\lambda_t^v$. It holds
$$
\sum_{i\in \mathcal{A}\setminus \{I_t(v)\}} p_t^v(i)\frac{1}{\left(1+\sqrt{p_t(i)}\gamma\right)^2} + p_t^v(I_t(v)) \frac{1}{\left(1+\sqrt{p_t^v(I_t(v))}(\eta\hat{\ell}_t^{v,obs}(I_t(v))+\gamma)\right)^2}=1.
$$

By Jensen's inequality,
$$
\frac{1}{\left(1 + \sum_{i\in \mathcal{A}\setminus \{I_t(v)\}}p_t^v(i)\sqrt{p_t^v(i)}\gamma + p_t^v(I_t(v))\sqrt{p_t^v(I_t(v))}(\eta \hat{\ell}_t^{v,obs}(I_t(v))+\gamma)\right)^2}\leq 1.
$$
This is equivalent to
$$
\sum_{i\in \mathcal{A}\setminus \{I_t(v)\}}p_t^v(i)\sqrt{p_t^v(i)}\gamma + p_t^v(I_t(v))\sqrt{p_t^v(I_t(v))}(\eta(v) \hat{\ell}_t^{v,obs}(I_t(v))+\gamma)\geq 0
$$
from which it follows
$$
\gamma \geq -\eta(v)\frac{p^v_t\left(I_t(v)\right)^{3/2}}{\sum_{i\in\mathcal{A}}p_t^v(i)^{3/2}}\hat{\ell}_t^{v,obs}(I_t(v))
\geq -\eta(v)\frac{p^v_t\left(I_t(v)\right)^{3/2}}{q^v_{t-d}(I_t(v))\sum_{i\in\mathcal{A}}p_t^v(i)^{3/2}}
$$
where the second inequality comes from $\ell_{t-d}(I_t(v))\mathbb{I}\{I_t(v)\in\mathcal{J}_{t-d}(v)\}\leq 1$.

Here we show that for any $\delta>1$, there exists a learning rate $\eta(v)>0$ such that 
$$
p_{t}^v(i) \leq \delta p_{t-1}^v(i) \quad \text{for all } i\in\mathcal{A}
$$
for all $t=2,\dots,T$ by induction over $t$.

\textit{Base step:}
it is trivial to see that the above inequalities hold for $t=2, \dots, d+1$
by noticing that
$$
\hat{\ell}^{v,obs}_t(i) = 0 \quad \text{for all } 1\leq t\leq d
$$
from which, combined with (\ref{equ:pt}) and (\ref{equ:lambdat}), it follows that 
$$
p^v_1(i) = \cdots = p^v_{d}(i) = p^v_{d+1}(i) \quad \text{for all } i\in \mathcal{A}.
$$
Hence, for $\delta>1$,
$$
p_{t}^v(i) = p_{t-1}^v(i) \leq \delta p_{t-1}^v(i) \quad \text{for all } i\in\mathcal{A}
$$
holds for $t=2,\dots,d+1$.

\textit{Induction step:} suppose that $p_{s}^v(i)\leq \delta p_{s-1}^v(i)$ holds for all $i\in\mathcal{A} $ and for all $s=2,\dots,t$ where $1+d\leq t \leq T-1$,  
then we have
$$
p^v_{t}(i)\leq \delta p^v_{t-1}(i) \leq \cdots \leq \delta^d p^v_{t-d}(i).
$$
Specifically, we have $p^v_{t}(I_t(v))\leq \delta^d p^v_{t-d}(I_t(v))$.

From this, it follows
$$
\frac{p^v_t\left(I_t(v)\right)^{3/2}}{q^v_{t-d}(I_t(v))\sum_{i\in\mathcal{A}}p_t^v(i)^{3/2}} \leq \delta^{3d/2}\frac{p^v_{t-d}\left(I_t(v)\right)^{3/2}}{q^v_{t-d}(I_t(v))\sum_{i\in\mathcal{A}}p_t^v(i)^{3/2}}
\leq \delta^{3d/2}\frac{\sqrt{p^v_{t-d}\left(I_t(v)\right)}}{\sum_{i\in\mathcal{A}}p_t^v(i)^{3/2}}
$$
where the second inequality comes from $p^v_{t-d}(I_t(v))\leq q^v_{t-d}(I_t(v))$.

This implies
$$
\sqrt{p_t^v(i)}\gamma \geq -\eta(v) \delta^{3d/2}\frac{\sqrt{p^v_{t-d}\left(I_t(v)\right)} \sqrt{p_t^v(i)}}{\sum_{i\in\mathcal{A}}p_t^v(i)^{3/2}}  \geq -\eta(v) \delta^{3d/2}\frac{1}{\sum_{i\in\mathcal{A}}p_t^v(i)^{3/2}}  \geq -\eta(v) \delta^{3d/2}\sqrt{K} 
$$
where the last inequality is because $\sum_{j=1}^K p_t^v(j)^{3/2} = K \sum_{j=1}^K (1/K)p_t^v(j)^{3/2}\geq K (\sum_{j=1}^K (1/K) p_t^v(j))^{3/2} = 1/\sqrt{K}$.
From (\ref{equ:pt}), it follows that for every $i\neq I_t(v)$,
$$
p_{t+1}^v(i) = \frac{1}{\left(1+\sqrt{p_t^v(i)}\gamma\right)^2}p_t^v(i)
\leq \frac{1}{\left(1-\eta(v) \delta^{3d/2}\sqrt{K} \right)^2} p_t^v(i) \leq \delta p^v_t(i)
$$
whenever $\eta(v)\leq (1-1/\sqrt{\delta})/\left(\delta^{3d/2}\sqrt{K}\right)$.

Clearly, since $\delta>1$, we have
$$
p^v_{t+1}(I_t(v))\leq p^v_t(I_t(v)) \leq \delta p^v_t(I_t(v)).
$$
Hence,
$$
p_{s}^v(i) \leq \delta p_{s-1}^v(i) \quad \text{for all } i\in\mathcal{A}
$$
holds for $s=t+1$.

Now we have shown both the base step and induction step.

From the induction, for any $\delta>1$,
$$
p_{t+1}^v(i) \leq \delta p_{t}^v(i)
$$
whenever $\eta(v)\leq (1-1/\sqrt{\delta})/\left(\delta^{3d/2}\sqrt{K}\right)$.

From (\ref{equ:pt}) and $\gamma\leq 0$, for all $i\in \mathcal{A}$,
$$
p_t^v(i) \leq \left(1+\sqrt{p_t^v(i)}(\eta(v)\hat{\ell}_t^{v,obs}(i))\right)^2 p_{t+1}^v(i).
$$
Now note that 
$$
\left(1+\sqrt{p_t^v(i)}(\eta(v)\hat{\ell}_t^{v,obs}(i))\right)^2
\leq \left(1+\eta(v)\hat{\ell}_t^{v,obs}(i)\right)^2 
\leq \frac{1+\eta(v)\hat{\ell}_t^{v,obs}(i)}{1-\eta(v)\hat{\ell}_t^{v,obs}(i)}
$$
from which it follows that
$$
p_t^v(i) \leq \frac{1+\eta(v)\hat{\ell}_t^{v,obs}(i)}{1-\eta(v)\hat{\ell}_t^{v,obs}(i)} p_{t+1}^v(i).
$$
This is equivalent to
$$
\left(1-\eta(v)\hat{\ell}_t^{v,obs}(i)\right)p_t^v(i) \leq \left(1+\eta(v)\hat{\ell}_t^{v,obs}(i)\right) p_{t+1}^v(i) = p_{t+1}^v(i) + \eta(v)\hat{\ell}_t^{v,obs}(i) p_{t+1}^v(i).
$$
It follows that when $\eta(v)\leq (1-1/\sqrt{\delta})/(\delta^{3d/2}\sqrt{K})$,
$$
\left(1-\eta(v)\hat{\ell}_t^{v,obs}(i)\right)p_t^v(i) \leq p_{t+1}^v(i) + \eta(v)\hat{\ell}_t^{v,obs}(i) p_{t+1}^v(i) \leq p_{t+1}^v(i) + \delta\eta(v)\hat{\ell}_t^{v,obs}(i) p_{t}^v(i).
$$
Rearranging the terms, we have
$$
\left(1-(1+\delta)\eta(v)\hat{\ell}_t^{v,obs}(i)\right)p_t^v(i) \leq p_{t+1}^v(i).
$$
Hence, we have 
$$
\left(1-(1+\delta)\eta(v)\hat{\ell}_t^{v,obs}(i)\right)p_t^v(i) \leq p_{t+1}^v(i) \leq \delta p_{t}^v(i)
$$
whenever $\eta(v)\leq (1-1/\sqrt{\delta})/(\delta^{3d/2}\sqrt{K})$.
Letting $\delta=2$ completes the proof.

\subsubsection{Proof of Theorem~3.1}
\label{sec::center-based}
Recall from the proof of Lemma~3.3 that for any $\delta>1$, when $\eta(v)\leq (1-1/\sqrt{\delta})/(\delta^{3d/2}\sqrt{K})$, then it holds
$$
\left(1-(1+\delta)\eta(v)\hat{\ell}_t^{v,obs}(i)\right)p_t^v(i) \leq p_{t+1}^v(i) \leq \delta p_{t}^v(i).
$$
Let $\delta=1+1/d$. Then, we have that for each center agent $c\in \mathcal{C}$,
\begin{equation}
\label{eq::p-center-non-center}
p_{t}^c(i) \leq (1+1/d) p_{t-1}^c(i) \leq  \cdots \leq (1+1/d)^d p_{t-d}^c(i) \leq 3p_{t-d}^c(i)
\end{equation}
and
\begin{equation}
\label{eq:regret-center-non-center}
p_t^c(i)
\leq p_{t+1}^c(i) + (1+\delta)\eta(c)p_t^c(i)\hat{\ell}_t^{c,obs}(i) 
= p_{t+1}^c(i) + (2+1/d)\eta(c)p_t^c(i)\hat{\ell}_t^{c,obs}(i)
\end{equation}
when 
\begin{equation*}
    \eta(c)  \leq \frac{1-\sqrt{d/(d+1)}}{(1+1/d)^{3d/2}\sqrt{K}}.
\end{equation*}

From Algorithm~1 and (\ref{eq::p-center-non-center}), it follows that
\begin{eqnarray*}
    q_t^c(i)
    &=& 1-\prod_{v\in \mathcal{N}(c)}(1-p_t^v(i)) \nonumber\\
    &=& 1-\prod_{v\in \mathcal{N}(c)}(1-p_{t-d}^c(i)) \nonumber\\
    &\geq & 1-\left(1-\frac{1}{3}p_t^c(i)\right)^{|\mathcal{N}(c)|} \nonumber\\
    &\geq & 1-\exp\left(-\frac{1}{3}p_t^c(i)|\mathcal{N}(c)|\right) \nonumber\\
    &\geq & (1-e^{-1})\min\left\{1, \frac{1}{3}p^c_t(i)|\mathcal{N}(c)|\right\} \nonumber \\
\end{eqnarray*}
from which it follows
\begin{equation}\label{eq:qc}
    \frac{1}{q^c_t(i)} \leq \frac{1}{(1-e^{-1})\min\{1, \frac{1}{3}p^c_t(i)|\mathcal{N}(c)|\}}
\leq 2 + \frac{6}{|\mathcal{N}(c)|p^c_t(i)}.
\end{equation}

For the Tsallis entropy regularizer (1) in Algorithm~1, we have
\begin{equation}
f_t^{\prime\prime}(x) = \frac{1}{2\eta(c) x^{3/2}}.
\label{equ:ft}
\end{equation}
From Lemma~3.2 and (\ref{equ:ft}), we have that the individual regret of center $c$ is bounded as
\begin{eqnarray}
    R_{T}^c
    &\leq & M + \frac{1}{2}  \mathbb{E} \left[\sum_{t=1}^T  \sum_{i\in \mathcal{A}} \frac{1}{q_t^c(i) f_t^{\prime\prime}(p_t^c(i))}\right] +  d\mathbb{E} \left[\sum_{t=1}^T \sum_{i\in \mathcal{A}} \frac{1}{f_t^{\prime\prime}(p_t^c(i))} \right]\nonumber\\
    &\leq & 2\frac{1}{\eta(c)} \sqrt{K} + \eta(c)  \mathbb{E} \left[\sum_{t=1}^T\sum_{i\in \mathcal{A}} \frac{p_t^c(i)^{\frac{3}{2}}}{q_t^c(i)} \right]+ 2d\eta(c)  \mathbb{E} \left[\sum_{t=1}^T \sum_{i\in \mathcal{A}} p_t^c(i)^{\frac{3}{2}}\right].\label{equ:rtc}
\end{eqnarray}
Since $p^c_t(i) \leq 1$, we have
\begin{equation}
    \label{eq:p-delay-term}
    \sum_{t=1}^T \sum_{i\in \mathcal{A}} p_t^c(i)^{\frac{3}{2}} \leq \sum_{t=1}^T \sum_{i\in \mathcal{A}} p_t^c(i) = T.
\end{equation}

Let us define $M:\mathcal{V}\rightarrow \mathbf{R}$ as follows
\begin{equation}
    \label{eq::mass}
        M(c) = \min \{|\mathcal{N}(c)|, K\}, \hbox{ for } c\in \mathcal{C} \hbox{ and }
        M(v) = e^{-\frac{1}{6} d(v)}M(C(v)) \hbox{ for } v\in \mathcal{V}\setminus \mathcal{C}.
\end{equation}

Now, note
\begin{eqnarray}
\mathbb{E} \left[\sum_{t=1}^T\sum_{i\in \mathcal{A}} \frac{p_t^c(i)^{\frac{3}{2}}}{q_t^c(i)} \right]
& \leq & \mathbb{E} \left[\sum_{t=1}^T\sum_{i\in \mathcal{A}} p_t^c(i)^{\frac{3}{2}}\left(2 + \frac{6}{|\mathcal{N}(c)|p^c_t(i)}\right) \right] \nonumber\\
& \leq & 
2T + \frac{6}{M(c)} \mathbb{E} \left[\sum_{t=1}^T\sum_{i\in \mathcal{A}} \sqrt{p_t^c(i)}\right] \nonumber\\
& \leq &
2T + \frac{6T\sqrt{K}}{M(c)}
\label{equ:pqsum}
\end{eqnarray}
where the first inequality is by (\ref{eq:qc}), the second inequality is by (\ref{eq:p-delay-term}) and (\ref{eq::mass}), and the last inequality is by Cauchy-Schwartz inequality.

From (\ref{equ:rtc}), (\ref{eq:p-delay-term}), and (\ref{equ:pqsum}), we have
$$
R_{T}^c
\leq 
2\frac{1}{\eta(c)} \sqrt{K} + \eta(c)\frac{6T\sqrt{K}}{M(c)} + 2(d+1)\eta(c)T.
$$
By letting $\eta(c) = \sqrt{M(c)/(3T)}$, it follows that
\begin{equation}
    \label{eq::center-individual}
    \begin{aligned}
    R_{T}^c
    &\leq 4\sqrt{3}\sqrt{\frac{K T}{M(c)} } + \frac{2(d+1)}{\sqrt{3}}\sqrt{M(c)T}.
    \end{aligned}
\end{equation}
Note that when
$$T\geq 36(d+1)^2K\max_{c}M(c) > 9K\max_{c}M(c) \left(\frac{1+\sqrt{1-\frac{1}{d+1}}}{\frac{1}{d+1}}\right)^2 = \frac{9K\max_{c}M(c)}{\left(1-\sqrt{\frac{d}{d+1}}\right)^2},$$
it follows that 
$$
\eta(c)= \sqrt{\frac{M(c)}{3T}}\leq \frac{1-\sqrt{\frac{d}{d+1}}}{3^{3/2}\sqrt{K}} \leq \frac{1-\sqrt{\frac{d}{d+1}}}{(1+1/d)^{3d/2}\sqrt{K}}.
$$
Hence (\ref{eq::p-center-non-center}) and (\ref{eq:regret-center-non-center}) holds for every center $c$.

Now, note that for any non-center agent $v$, $p_t^v = p^{C(v)}_{t-d(v)d}$.
Then, from (\ref{eq:regret-center-non-center}), for all $t>d(v)d$,
\begin{eqnarray}
    p_t^v(i) 
    &=& p^{C(v)}_{t-d(v)d}(i) \nonumber \\
    &\leq& p^{C(v)}_{t-d(v)d+1}(i) + (2+1/d)\eta(C(v)) p^{C(v)}_{t-d(v)d}(i) \hat{\ell}^{C(v),obs}_{t-d(v)d}(i) \nonumber \\
    &\cdots& \nonumber\\
    &\leq& p^{C(v)}_{t}(i) + (2+1/d)\eta(C(v)) \sum_{s=1}^{d(v)d} p^{C(v)}_{t-s}(i) \hat{\ell}^{C(v),obs}_{t-s}(i). \label{eq::non-center-prob}
\end{eqnarray}

For every non-center agent $v$, the individual regret is bounded as 
\begin{eqnarray}
    R_T^v 
    &\leq & (d(v)+1)d + \mathbb{E}\left[ \sum_{t=(d(v)+1)d+1}^{T} \sum_{i\in \mathcal{A}} p_{t}^{v}(i) \ell_{t}(i)-\min _{i \in \mathcal{A}} \sum_{t=(d(v)+1)d+1}^{T} \ell_{t}(i)\right] \nonumber\\
    &\leq &  (d(v)+1)d + \mathbb{E}\left[\sum_{t=(d(v)+1)d+1}^{T}\sum_{i\in \mathcal{A}} p^{C(v)}_{t}(i) \ell_{t}(i)-\min _{i \in \mathcal{A}} \sum_{t=(d(v)+1)d+1}^{T} \ell_{t}(i)\right] \nonumber \\
    & & + (2+1/d)\eta(C(v)) \mathbb{E}\left[\sum_{t=(d(v)+1)d+1}^{T} \sum_{i\in \mathcal{A}} \sum_{s=1}^{d(v)d} p_{t-s}^{C(v)}(i) \hat{\ell}_{t-s}^{C(v), obs}(i)\ell_{t}(i)\right]  \nonumber \\
    &=& (d(v)+1)d+ \mathbb{E}\left[\sum_{t=(d(v)+1)d+1}^{T} \sum_{i\in \mathcal{A}} p^{C(v)}_{t}(i) \ell_{t}(i)-\min _{i \in \mathcal{A}} \sum_{t=(d(v)+1)d+1}^{T} \ell_{t}(i)\right] \nonumber \\
    & & + (2+1/d)\eta(C(v)) \mathbb{E}\left[\sum_{t=(d(v)+1)d+1}^{T} \sum_{i\in \mathcal{A}} \sum_{s=1}^{d(v)d} p_{t-s}^{C(v)}(i) \hat{\ell}_{t-s-d}^{C(v)}(i) \ell_{t}(i)\right]  \label{eq::individual-regret} 
\end{eqnarray}
where the second inequality follows from (\ref{eq::non-center-prob}) and the equality uses $\hat{\ell}_{t-d}^{v} = \hat{\ell}_{t}^{v, obs}$ when $t>d$.

From Lemma~\ref{lm:conditional_exp}, we have
\begin{eqnarray*}
&& \mathbb{E}\left[\sum_{t=(d(v)+1)d+1}^{T} \sum_{i\in\mathcal{A}} \sum_{s=1}^{d(v)d} p_{t-s}^{C(v)}(i) \hat{\ell}_{t-s-d}^{C(v)}(i) \ell_{t}(i)\right]\\
& \leq &
\mathbb{E}\left[\sum_{t=(d(v)+1)d+1}^{T}\sum_{i\in \mathcal{A}} \sum_{s=1}^{d(v)d} p_{t-s}^{C(v)}(i)\mathbb{E}_{t-s-d}\left[ \hat{\ell}_{t-s-d}^{C(v)}(i) \right]\right]
\\
& \leq & d(v)dT.
\end{eqnarray*}

Hence, it follows that
$$
R_T^v \leq R_{T}^{C(v)}+(d(v)+1)d+(2+1/d) \eta(C(v)) d(v)dT.
$$

Combining with Eq.~(\ref{eq::center-individual}), we have
$$
    \label{eq::non-center-individual}
    \begin{aligned}
    R_T^v
    &\leq 4\sqrt{3}\sqrt{\frac{K T}{M(C(v))} } + \frac{2(d+1)}{\sqrt{3}}\sqrt{M(C(v))T} + (d(v)+1)d+(2+1/d)\eta(C(v)) d(v)dT  \\
    &= 4\sqrt{3}\sqrt{\frac{K T}{M(C(v))} } + \frac{2(d+1)}{\sqrt{3}}\sqrt{M(C(v))T} + (d(v)+1)d+(2d+1)d(v)\sqrt{M(C(v))T/3} \\
    &\leq 4\sqrt{3}\sqrt{\frac{K T}{M(v)} } + \frac{2(d+1)}{\sqrt{3}}\sqrt{M(C(v))T} + (d(v)+1)d+(2d+1)d(v)\sqrt{M(C(v))T/3} \\
    &\leq 4\sqrt{3}\sqrt{\frac{K T}{M(v)} } + \frac{2(d+1)}{\sqrt{3}}\sqrt{M(C(v))T} + 6d\log(K)+2\sqrt{3}(2d+1)\log(K)\sqrt{M(C(v))T} \\
    \end{aligned}
$$
where the penultimate inequality follows from Eq.~(\ref{eq::mass}) and the last inequality follows from Lemma~7 in \cite{DBLP:conf/nips/Bar-OnM19}.

From Theorem~8 in \cite{DBLP:conf/nips/Bar-OnM19}, we have for all agent $v\in \mathcal{V}$
$$
M(v) \geq (1/e) \min \{|\mathcal{N}(v)|, K\}
$$
Hence, we have
\begin{equation*}
    \begin{aligned}
    R_T^v
    &\leq 
    4\sqrt{3}\sqrt{e}\sqrt{\frac{KT}{ \min \{|\mathcal{N}(v)|, K\}}}+  \frac{2(d+1)}{\sqrt{3}}\sqrt{M(C(v))T}+ 6d\log(K) + 2\sqrt{3}(2d+1)\log(K)\sqrt{M(C(v))T}\\
    &\leq 
    12\sqrt{\frac{KT}{|\mathcal{N}(v)|}}+  \frac{2(d+1)}{\sqrt{3}}\sqrt{|\mathcal{N}(C(v))|T}+ 6d\log(K) + 2\sqrt{3}(2d+1)\log(K)\sqrt{|\mathcal{N}(C(v))|T}\\
    \end{aligned}
\end{equation*}
where the second inequality comes from the fact that $K\geq\max_v{|\mathcal{N}(v)|}$.

\subsubsection{Proof of Lemma~3.5}
\label{sec::upper-bound-lemma-2}

The proof follows by the following sequence of relations:

\begin{equation*}
    \begin{aligned}
    \sum_{i\in \mathcal{A}}\sum_{v \in \mathcal{V}} \frac{p^v(i)^{3/2}}{q^v(i)} &=  \sum_{i\in \mathcal{A}}\sum_{v \in \mathcal{V}} \left(\frac{p^v(i)}{q^v(i)} \cdot \sqrt{p^v(i)}\right) & \\ 
    &\leq \sum_{i\in \mathcal{A}}\sqrt{\sum_{v \in \mathcal{V}}\frac{p^v(i)^2}{q^v(i)^2}}\sqrt{\sum_{v \in \mathcal{V}} p^v(i)} & (\text{Cauchy-Schwarz inequality})\\
    &\leq \sum_{i\in \mathcal{A}}\sqrt{\sum_{v \in \mathcal{V}}\frac{p^v(i)}{q^v(i)}}\sqrt{\sum_{v \in \mathcal{V}} p^v(i)} & (q^v(i) \geq p^v(i))\\
    &\leq \sum_{i\in \mathcal{A}}\sqrt{\frac{1}{1-1/e}\left(\alpha(G)+\sum_{v \in \mathcal{V}} p^v(i)\right)}\sqrt{\sum_{v \in \mathcal{V}} p^v(i)} & (\text{Lemma~3 in \cite{DBLP:conf/alt/Cesa-BianchiCM20}}) \\
    &\leq \sqrt{\sum_{i\in \mathcal{A}} \frac{1}{1-1/e}\left(\alpha(G)+\sum_{v \in \mathcal{V}} p^v(i)\right)}
    \sqrt{\sum_{i\in \mathcal{A}} \sum_{v \in \mathcal{V}} p^v(i)} & (\text{Cauchy-Schwarz inequality})\\
    &= \sqrt{\frac{N}{1-1/e}\left(K\alpha(G)+N\right)}. & \\
    \end{aligned}
\end{equation*}

\subsubsection{Proof of Theorem~3.4}
\label{sec::upper-bound-theorem-1}

Note that 

$$
\max_{x \in \mathcal{P}_{K-1}}\{-F_{1}(x)\}+\sum_{t=2}^{T} \max_{x \in \mathcal{P}_{K-1}}\left\{F_{t-1}(x)-F_{t}(x)\right) = -F_{T}(\mathrm{e}_{[K]}/K) \\
    = 2\frac{1}{\eta_T} \sqrt{K} + \frac{1}{\zeta_T}\log(K),
$$
and
$$
f_t^{\prime\prime}(x) \geq \max\left\{\frac{1}{2\eta_t} \frac{1}{x^{3/2}}, \frac{1}{\zeta_t}\frac{1}{x}\right\}.
$$
Hence, from Lemma~3.5, we have
\begin{equation*}
    \begin{aligned}
    \frac{1}{2N}\mathbb{E}\left[\sum_{t=1}^T \sum_{i\in \mathcal{A}} \sum_{v\in \mathcal{V}} \frac{1}{q_t^v(i) f^{\prime\prime}_t(p_t^v(i))} \right]
    &\leq \frac{1}{N} \mathbb{E}\left[  \sum_{t=1}^T \eta_t\sum_{i\in \mathcal{A}} \sum_{v\in \mathcal{V}} \frac{p_t^v(i)^{3/2}}{q_t^v(i) } \right] 
    \\
    &\leq \sqrt{\frac{1}{1-1/e}\left(\frac{K}{N}\alpha(G)+1\right)} \sum_{t=1}^T\eta_t
    \end{aligned}
\end{equation*}
and
\begin{equation*}
    \begin{aligned}
    \frac{d}{N} \mathbb{E}\left[ \sum_{t=1}^T \sum_{i\in \mathcal{A}} \sum_{v\in \mathcal{V}} \frac{1}{f^{\prime\prime}_t(p_t^v(i))}  \right] 
    &\leq \frac{d}{N}  \mathbb{E}\left[ \sum_{t=1}^T \zeta_t \sum_{i\in \mathcal{A}} \sum_{v\in \mathcal{V}} p_t^v(i) \right] \\
    &= d\sum_{t=1}^T \zeta_t.
    \end{aligned}
\end{equation*}

Combining with Lemma~3.2, we have
\begin{equation*}
    \begin{aligned}
        R_{T} 
        &= \frac{1}{N} \sum_{v\in \mathcal{V}} R_T^v \\
        &\leq M + \frac{1}{2N} \mathbb{E}\left[ \sum_{t=1}^T  \sum_{i\in \mathcal{A}} \sum_{v\in \mathcal{V}} \frac{1}{q_t^v(i) f_t^{\prime\prime}(p_t^v(i))} \right]+ \frac{d}{N} \mathbb{E}\left[ \sum_{t=1}^T \sum_{i\in \mathcal{A}} \sum_{v\in \mathcal{V}} \frac{1}{f_t^{\prime\prime}(p_t^v(i))}  \right] \\
        &\leq  2\frac{1}{\eta_T}\sqrt{K} + \sqrt{\frac{1}{1-1/e}\left(\frac{K}{N}\alpha(G)+1\right)} \sum_{t=1}^T\eta_t + \frac{1}{\zeta_T}\log(K) + d\sum_{t=1}^T \zeta_t.
    \end{aligned}
\end{equation*}
Plugging 
$\eta_t = (1/(1-1/e)) (\alpha(G)/ N + 1/K)^{-1/4} \sqrt{2/T}$ and $\zeta_t = \sqrt{\log(K) / (dt)}$
into the inequality above, we complete the proof.

\subsection{Lower bounds}

\subsubsection{Proof of Theorem~4.1}
\label{sec::lower-bound}

In each round $t$, every agent $v\in \mathcal{V}$ receives $b_t(v) = O(|\mathcal{N}(v)|)$ bits since its neighboring agents can choose at most $\mathcal{N}(v)$ distinct actions. By Theorem~4 in \cite{DBLP:conf/nips/Shamir14}, there exists some distribution $\mathcal{D}$ over $[0,1]^K$ such that loss vectors $\ell_t\sim \mathcal{D}$ for all $t=1,2,\dots, T$ independently and $\min_{i\in \mathcal{A}} \mathbb{E}\left[\ell_t(I_t(v)) - \ell_t(i)\right] = \Omega\left(\min\{T, \sqrt{KT/|\mathcal{N}(v)|}\}\right)$
Hence, the worst-case individual regret of agent $v$ is
\begin{equation*}
    \begin{aligned}
    \sup_{\ell_1\ldots,\ell_T} R_T^v 
    &=  \sup_{\ell_1\ldots,\ell_T} \mathbb{E}\left[\sum_{t=1}^{T} \ell_{t}\left(I_{t}(v)\right)-\min _{i \in \mathcal{A}} \sum_{t=1}^{T} \ell_{t}(i)\right] \\
    &\geq \mathbb{E}_{\ell_t\sim \mathcal{D}}\left[ \mathbb{E}\left[\sum_{t=1}^{T} \ell_{t}\left(I_{t}(v)\right) \right] \right] - \mathbb{E}_{\ell_t\sim \mathcal{D}}\left[\min _{i \in \mathcal{A}} \sum_{t=1}^{T} \ell_{t}(i)\right] \\
    &\geq \mathbb{E}_{\ell_t\sim \mathcal{D}}\left[ \mathbb{E}\left[\sum_{t=1}^{T} \ell_{t}\left(I_{t}(v)\right) \right] \right] - \min _{i \in \mathcal{A}}  \mathbb{E}_{\ell_t\sim \mathcal{D}}\left[\sum_{t=1}^{T} \ell_{t}(i)\right] \\
    &= \max_{i\in\mathcal{A}} \mathbb{E}\left[\sum_{t=1}^{T} \ell_{t}\left(I_{t}(v)\right) -  \sum_{t=1}^{T} \ell_{t}(i) \right]\\
    &\geq \min_{i\in\mathcal{A}} \mathbb{E}\left[\sum_{t=1}^{T} \ell_{t}\left(I_{t}(v)\right) -  \sum_{t=1}^{T} \ell_{t}(i) \right]\\
    &\geq \Omega\left(\min\{T, \sqrt{KT/|\mathcal{N}(v)|}\}\right).
    \end{aligned}
\end{equation*}
Note that with delay $d$, the individual regret of the agent $v$ can not be smaller than the regret of the agent $v$ with access to full information of $\ell_1,\dots, \ell_{t-d}$ at each round $t$.
Following the argument in the proof of Corollary~15 in \cite{DBLP:journals/jmlr/Cesa-BianchiGM19}, the worst-case regret of a single-agent with delayed full information is $\Omega(\sqrt{dT\log(K)})$.
Hence,
\begin{equation*}
    \sup_{\ell_1,\ldots,\ell_T} R_T^v = \Omega\left(\max\left\{ \min \left\{T, \sqrt{\frac{KT}{|\mathcal{N}(v)|}}\right\}, \sqrt{dT \log(K)}\right\} \right).
\end{equation*}
We complete the proof by noting that $R_T = \frac{1}{N}\sum_{v\in \mathcal{V}} R_T^v$.

\subsection{Discussion on the regret bounds}

\subsubsection{CFTRL and the center-based Exp3 when $d=1$}
\label{sec::regret-comparison}
As noted in Theorem~3.1 with $d=1$ and \cite{DBLP:conf/nips/Bar-OnM19}, when $K\geq \max_{v\in V}|\mathcal{N}(v)|$ and $T\geq \max\{144K \max_{c\in \mathcal{C}}|\mathcal{N}(c)|, K^2\log(K)\}$, the individual regret of CFTRL is 
$$
O\left(\sqrt{\left(\frac{1}{|\mathcal{N}(v)|}+\frac{|\mathcal{N}(C(v))|}{K}\log(K)^2\right)KT}\right),
$$ and the individual regret of the center-based Exp3 is
$$
\tilde{O}\left(\sqrt{\left(1+\frac{K}{\mathcal{N}(v)}\right)T}\right) = O\left(\sqrt{\left(\frac{1}{|\mathcal{N}(v)|}+\frac{1}{K}\right)\log(K)KT}\right)
$$
If we want the individual regret of CFTRL lower than the center-based Exp3, it suffices to show
$$
\frac{1}{|\mathcal{N}(v)|} + \frac{|\mathcal{N}(C(v))|}{K}\log(K)^2 \leq \left(\frac{1}{|\mathcal{N}(v)|}+\frac{1}{K}\right)\log(K) 
$$
which is equivalent to
$$
\frac{|\mathcal{N}(C(v))|}{K}\log(K)^2 - \frac{\log(K)}{K} -  \leq \frac{\log(K)-1}{|\mathcal{N}(v)|}  
$$
Note that for $K\geq 9$, we have
$$
\log(K) - 1\geq \frac{1}{2}\log(K)
$$
It is suffices to show that
$$
\frac{|\mathcal{N}(C(v))|}{K}\log(K)^2 \leq \frac{\log(K)}{2|\mathcal{N}(v)|}  
$$
which is equivalent to
$$
\frac{K}{\log(K)} \geq 2 |\mathcal{N}(v)||\mathcal{N}(C(v))|.
$$

\subsubsection{CFTRL and the lower bound when $d=1$}
\label{sec::optimal-bound-condition}
As noted in Theorem~4.1, for $T\geq K/|\mathcal{N}(v)|$, the individual regret lower bound is 
$$
\Omega\left(\sqrt{\left(\frac{1}{|\mathcal{N}(v)|}+\frac{1}{K}\log(K)\right)KT}\right).
$$
From Theorem~3.1 with $d=1$, when  $K\geq \max_{v\in V}|\mathcal{N}(v)|$ and $T\geq 144K \max_{c\in \mathcal{C}}|\mathcal{N}(c)|$, the individual regret of CFTRL is 
$$
O\left(\sqrt{\left(\frac{1}{|\mathcal{N}(v)|}+\frac{|\mathcal{N}(C(v))|}{K}\log(K)^2\right)KT}\right).
$$
Hence when $K\geq \max_{v\in V}|\mathcal{N}(v)|$ and $T\geq \max\{144K \max_{c\in \mathcal{C}}|\mathcal{N}(c)|, K/|\mathcal{N}(v)|\}$,
the ratio between the individual regret of CFTRL and the individual regret lower bound is 
$$
O\left( \sqrt{\frac{\frac{1}{|\mathcal{N}(v)|}+\frac{|\mathcal{N}(C(v))|}{K}\log(K)^2}{\frac{1}{|\mathcal{N}(v)|}+\frac{1}{K}\log(K)}} \right).
$$
Note that 
$$
\sqrt{\frac{\frac{1}{|\mathcal{N}(v)|}+\frac{|\mathcal{N}(C(v))|}{K}\log(K)^2}{\frac{1}{|\mathcal{N}(v)|}+\frac{1}{K}\log(K)}}
\leq 1 + \sqrt{\frac{|\mathcal{N}(C(v))|\log(K)-1}{\frac{K}{|\mathcal{N}(v)|\log(K)}+1}}
\leq 1 + \sqrt{\frac{|\mathcal{N}(C(v))||\mathcal{N}(v)|\log(K)^2}{K}}.
$$
When $|\mathcal{N}(C(v))||\mathcal{N}(v)|\log(K)^2/K = O(1)$, the individual regret upper bound of CFTRL matches the lower bound up to a constant factor.

\subsection{Numerical experiments}
\label{sec::numerical-experiments-app}

All the experiments are run on a desktop with AMD Ryzen 5 2600 Six-Core Processor and 16GB memory.
Each experiment took less than 6 hours to finish. 

The code is written in Python and uses Numpy package \citep{harris2020array} and NetworkX package \citep{hagberg2008exploring} for numerical calculation and graph operations.
The Numpy package and the NetworkX package are distributed with the BSD license.

\end{document}